\pdfoutput=1

\documentclass[11pt]{article}

\usepackage{acl}

\usepackage{times}
\usepackage{latexsym}

\usepackage[T1]{fontenc}

\usepackage[utf8]{inputenc}

\usepackage{microtype}

\usepackage{inconsolata}

\newcommand{\minute}[1]{\textcolor{black}{#1}}
\newcommand{\rparagraph}[1]{\vspace{1mm}\noindent\textbf{#1.}}
\newcommand{\rrparagraph}[1]{\vspace{0.4mm}\noindent\textit{#1.}}
\newcommand{\sparagraph}[1]{\vspace{0.0mm}\noindent\textbf{#1.}}
\newcommand{\ttr}{{\texttt{T-Train}}\xspace}
\newcommand{\tts}{{\texttt{T-Test}}\xspace}
\newcommand{\ttt}{{\texttt{RTT}}\xspace}

\newcommand{\source}{\texttt{SRC}\xspace} 
\newcommand{\target}{\texttt{TRG}\xspace} 
\newcommand{\hr}{\texttt{HR}\xspace} 
\newcommand{\rnd}{\texttt{RT}\xspace}
\newcommand{\rndm}{\texttt{M-RT}\xspace}
\newcommand{\tm}{\texttt{M-TRG}\xspace} 
\newcommand{\jointt}{\texttt{TRG+SRC}\xspace} 
\newcommand{\jointtm}{\texttt{M-TRG+SRC}\xspace} 
\newcommand{\jointth}{\texttt{TRG+SRC+HR}\xspace} 
\newcommand{\jointtmh}{\texttt{M-TRG+SRC+HR}\xspace} 
\newcommand{\jointh}{\texttt{SRC+HR}\xspace} 
\newcommand{\rndt}{\texttt{RT+SRC}\xspace} 
\newcommand{\rndtm}{\texttt{M-RT+SRC}\xspace} 
\newcommand{\rndens}{\texttt{M-RT-Ens}\xspace}
\newcommand{\rndsens}{\texttt{M-RT-Ens-SRC}\xspace} 
\newcommand{\rndhens}{\texttt{M-RT-Ens-HR}\xspace} 
\newcommand{\seqt}{\texttt{SRC}$\rightarrow$\texttt{TRG}\xspace} 
\newcommand{\seqtm}{\texttt{SRC}$\rightarrow$\texttt{M-TRG}\xspace} 

\newcommand{\src}{\textit{Val-Src}\xspace}
\newcommand{\mtsrc}{\textit{Val-MT-Trg}\xspace}
\newcommand{\oracle}{\textit{Val-Trg}\xspace}
\newcommand{\srct}{Val-Src\xspace}
\newcommand{\mtsrct}{Val-MT-Trg\xspace}
\newcommand{\oraclet}{Val-Trg\xspace}

\usepackage{multirow}
\usepackage{adjustbox}
\usepackage{booktabs}
\usepackage{tabularx}
\usepackage{verbatim}
\usepackage{hyperref}
\usepackage{xspace}
\usepackage{caption}
\usepackage{subcaption}
\usepackage{amssymb}

\title{To Translate or Not to Translate: A Systematic Investigation of Translation-Based Cross-Lingual Transfer to Low-Resource Languages}

\author{Benedikt Ebing and Goran Glava\v{s} \\
  University of W\"{u}rzburg \\ Center for Artificial Intelligence and Data Science (CAIDAS) \\
  \texttt{\{benedikt.ebing, goran.glavas\}@uni-wuerzburg.de} \\}

\begin{document}
\maketitle

\begin{abstract}

Perfect machine translation (MT) would render cross-lingual transfer (XLT) by means of multilingual language models (mLMs) superfluous. Given, on the one hand, the large body of work on improving XLT with mLMs and, on the other hand, recent advances in massively multilingual MT, in this work, we systematically evaluate existing and propose new translation-based XLT approaches for transfer to low-resource languages. We show that all translation-based approaches dramatically outperform zero-shot XLT with mLMs---with the combination of round-trip translation of the source-language training data and the translation of the target-language test instances at inference---being generally the most effective. We next show that one can obtain further empirical gains by adding reliable translations to other high-resource languages to the training data. Moreover, we propose an effective translation-based XLT strategy even for languages not supported by the MT system. Finally, we show that model selection for XLT based on target-language validation data obtained with MT outperforms model selection based on the source-language data. We believe our findings warrant a broader inclusion of more robust translation-based baselines in XLT research.     

\end{abstract}
\section{Introduction}
Multilingual language models (mLMs) like mBERT \cite{devlin-etal-2019-bert}, XLM-R \cite{conneau-etal-2020-unsupervised}, or mT5 \cite{xue-etal-2021-mt5} have become the backbone of multilingual NLP. Their multilingual pretraining and the consequent ability to encode texts from a wide range of languages make them suitable for cross-lingual transfer (XLT) for downstream NLP tasks: fine-tuned on available task-specific data in high-resource languages, they can be used to make predictions for languages that lack task-specific (training) data. Their effectiveness as vehicles of both zero-shot (no task-specific training instances in the target language, ZS-XLT) and few-shot XLT (few training instances in the target language, FS-XLT) has been documented for a plethora of tasks and languages \cite{wu-dredze-2019-beto, wang2019cross,lauscher-etal-2020-zero, schmidt-etal-2022-dont}. 
Cross-lingual transfer with mLMs, however, yields poor performance for low-resource target languages that are (i) un(der)represented in the pretraining corpora, especially if they are additionally (ii) linguistically distant from the source language \cite{lauscher-etal-2020-zero,adelani-etal-2022-masakhaner,ebrahimi-etal-2022-americasnli}.

Recent years have witnessed a large body of work that focused on improving XLT, in particular for low-resource target languages. First, a multitude of new multilingual benchmarks have been introduced, aiming to either evaluate XLT with mLMs on sets of linguistically diverse languages \cite{clark2020tydi,ponti-etal-2020-xcopa, ruder-etal-2021-xtreme} or on groups of related low-resource languages from underrepresented language families (i.e., families without any high-resource language) and/or geographies \cite[\textit{inter alia}]{adelani-etal-2021-masakhaner, adelani-etal-2022-masakhaner, ebrahimi-etal-2022-americasnli, aggarwal-etal-2022-indicxnli, muhammad-etal-2022-naijasenti, armstrong-etal-2022-jampatoisnli, winata-etal-2023-nusax}. Second, a diverse set of methodological proposals have been introduced, ranging from (i) attempts to better align mLMs' representation subspaces of languages \cite[inter alia]{wu-dredze-2020-explicit,hu-etal-2021-explicit,yang2022enhancing} over (ii) those that increase the representational capacity for underrepresented languages, typically via additional post-hoc language-specific language modeling training \cite[\textit{inter alia}]{pfeiffer-etal-2020-mad,pfeiffer-etal-2022-lifting,ansell-etal-2022-composable,parovic-etal-2022-bad} to (iii) various FS-XLT proposals that seek to maximally exploit small sets of task-specific target language instances \cite[\textit{inter alia}]{hedderich-etal-2020-transfer,lauscher-etal-2020-zero,zhao-etal-2021-closer,schmidt-etal-2022-dont}.   

Much of the above work rendered \textit{translation-based XLT strategies}---in which an MT model is employed to either translate the source-language training data into the target language before training (referred to as \textit{translate-train}) or translate the target-language instances to the source language before inference (\textit{translate-test})---competitive w.r.t. mLM-based transfer \cite{pmlr-v119-hu20b, ruder-etal-2021-xtreme, ebrahimi-etal-2022-americasnli, aggarwal-etal-2022-indicxnli}. 
\textit{Sporadically}, however, MT has been leveraged for more elaborate translation-based strategies---e.g., translating source-language training data to multiple (related) target languages \cite{pmlr-v119-hu20b}, combining the translated target-language training data with the original source-language training data \cite{chen-etal-2023-frustratingly}, or using monolingual English LMs instead of mLMs for translate-test \cite{artetxe-etal-2020-translation,artetxe2023revisiting}---complicating the selection of translation-based baselines in XLT research. In fact, the most recent evidence \cite{artetxe2023revisiting} suggests that the potential of translation-based XLT has been underestimated due to the selection of suboptimal translation strategies. What is more, much of the work on low-resource XLT completely disregards translation-based baselines, arguing \textit{a priori}, without empirical confirmation, that (1) due to the lack of parallel data, MT models for low-resource languages exhibit poor performance, which directly caps the potential of translation-based XLT and/or (2) their evaluation encompasses target languages that are unsupported by (state-of-the-art, commercial) MT systems.  

Two recent developments, however, warrant a systematic (re-)evaluation of translation-based XLT for low-resource languages: (i) the availability of open massively multilingual MT models that not only support an increasingly large set of languages \cite{tiedemann-thottingal-2020-opus,liu2020multilingual, fan2021beyond,nllbteam2022language,kudugunta2023madlad}, but also yield meaningful translations even for the smallest of those languages; and  
(ii) recent proposals of novel translation-based XLT strategies that have been largely uninvestigated in XLT to truly low-resource languages \cite{pmlr-v119-hu20b,chen-etal-2023-frustratingly,artetxe2023revisiting}. 

\rparagraph{Contributions} In this work, we contribute to the body of translation-based XLT in light of these recent advances, focusing explicitly on low-resource target languages: 
\textbf{1)} we offer a systematic comparison of different translation-based XLT strategies on three established benchmarks for sequence- and token-level classification, encompassing in total 40 different low-resource languages; \textbf{2)} Motivated by the success of multi-source training \cite{ruder2017overview, ansell-etal-2021-mad-g} and ensembling \cite{oh-etal-2022-synergy}, as well as the high quality of MT between high-resource languages, we propose two novel strategies that integrate translations from the source data to three diverse high-resource languages (Turkish, Russian, and Chinese); we find that integrating translations to other high-resource languages substantially improves performance for sequence-level classification tasks; 
\textbf{3)} We propose a simple and effective translation-based XLT approach for languages \textit{not covered} by the MT models in which we translate from/to the linguistically closest supported language, demonstrating substantial gains over ZS-XLT with mLMs; 
\textbf{4)} We introduce a translation-based \textit{model selection} in which the optimal model checkpoint is selected based on performance on the validation data automatically translated to the target language; we show that this results in better performance than model selection based on source-language validation data. 
\textbf{5)} Finally, we run several ablations, offering insights into the impact of lower-level design decisions---such as the MT decoding strategy or joint vs. sequential fine-tuning---on translation-based XLT.

\section{Translation-Based Strategies}

\begin{figure*}
\centering
\includegraphics[width=\linewidth]{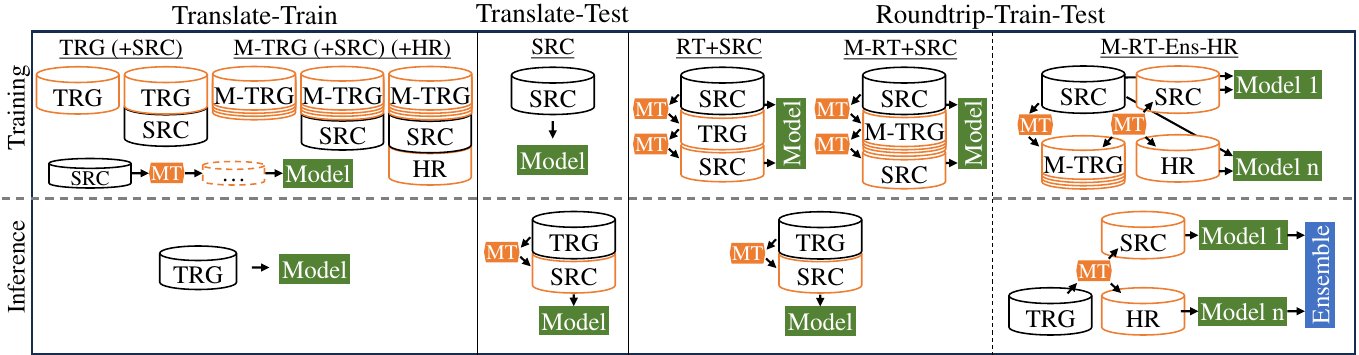} 
\caption{Schematic overview of translation-based XLT methods. Clean source or target language data is indicated in black, while noisy translated data is shown in orange.}
\label{fig:transXLT}
\end{figure*}

Most of the existing XLT work evaluates only the most straightforward \textit{translate-train} (\ttr) and \textit{translate-test} (\tts) baselines. The former assumes the translation of the training data, available in some high-resource language (almost always English), to the target language in which inference is performed. The latter trains on the clean source-language data but, at inference time, translates the target language instances to the source language before making predictions. More recent works \cite{oh-etal-2022-synergy,artetxe2023revisiting} propose a combination of the two, which we dub \textit{roundtrip-train-test} (\ttt), where the source-language training data is round-trip translated (i.e., to the target language and then back) so that the translated test data at inference time better matches the training distribution, reflecting the idiosyncrasies of the same MT model. In what follows, we describe the variants of \ttr, \tts, and \ttt that we evaluate. Figure \ref{fig:transXLT} concisely illustrates all MT-based approaches under evaluation.  

\subsection{Translate-Train (\ttr)}

\sparagraph{Target (\target)} This is the standard \ttr where the source-language training data is translated into one particular target language. The mLM is then fine-tuned on the automatically translated (i.e., noisy) target-language training dataset. 

\rparagraph{Multi-Target (\tm)} This is a generalization of \ttr in which we translate the source-language training data into each language from a set of (presumably related) target languages: in our experiments, these are all languages of a particular benchmark dataset supported by the MT model, e.g., all AmericasNLI languages \cite{ebrahimi-etal-2022-americasnli}. We then fine-tune the mLM in a multi-source setup, i.e., on the concatenation of the training data translated to each of the target languages (per task).

\rparagraph{Keeping the Source-Language Data (+\source)} In this variant, we concatenate the noisy translated training dataset in the target language (or a set of target languages) with the original (i.e., clean) training data in the source language. We denote these variants \jointt (if we concatenate source language data to \target) and \jointtm (if we concatenate the source-language data to \tm).

\rparagraph{Adding Diverse High-Resource Languages (+\hr)} We additionally explore translating the source-language training data to a (small) set of linguistically diverse high-resource languages. The motivation for this is two-fold: (1) multilingual (i.e., multi-source) fine-tuning has been shown to bring benefits compared to monolingual (English-only) fine-tuning \cite{ansell-etal-2021-mad-g}; and (2) automatic translation from high-resource source language (i.e., English) to other high-resource languages (i.e., Chinese, Turkish, and Russian) is generally of much higher quality than translation to low-resource target languages (e.g., Guarani). 
Exploiting strong MT between high-resource languages will, under this assumption, allow us to obtain linguistically diverse yet high-quality training data, which should consequently lead to improvements in XLT to any low-resource language. We evaluate variants in which translations to high-resource languages are added to \jointt (i.e., \jointth) and \jointtm (i.e., \jointtmh).       

\subsection{Translate-Test (\tts)}

We evaluate the standard \tts baseline where the model is trained on the original source-language data and, at inference time, the target-language instances are translated to the source language before the source-language model makes the prediction. 

\subsection{Roundtrip-Train-Test (\ttt)}

\sparagraph{Round-Trip \ttr + \tts (\rnd)} Prior work suggested that the mismatch between high-quality training data and noisily translated evaluation data poses a challenge for the \tts approach \cite{artetxe-etal-2020-translation,oh-etal-2022-synergy,artetxe2023revisiting}. To overcome this shift in data distribution that the model is exposed to at test time, in \ttt, we also train on the noisy source-language data obtained via round-trip translation of the original clean source-language data to the target language and back. Similar to \ttr, we evaluate the variants of \ttt where the noisy source-language data is obtained via round-trip translation to a single target language (denoted with \rnd) and multiple target languages (\rndm) and, finally, concatenated to the original (i.e., clean) source-language data (\rndt, \rndtm).

\rparagraph{Model Ensembling for \ttt (\rndens)} Following our idea of exploiting other high-resource languages in translation-based XLT, we propose a novel \ttt variant in which we not only round-trip translate the source-language data into the target language and back into the source language but also translate the source data into the target language and then into different high-resource languages, other than the initial source language (e.g., Source$\rightarrow$Target$\rightarrow$Chinese). We apply this paradigm to the same three high-resource languages used for the \ttr-based approaches (i.e., Chinese, Russian, and Turkish). Here in ensembling, however, for each of these high-resource languages, we independently fine-tune an mLM instance on the round-trip translated data of that language, concatenated with the original source-language (i.e., English) data (e.g., for English as source and Chinese as the high-resource auxiliary, we concatenate the clean original English with the noisy Chinese data obtained via two-step translation). Finally, we ensemble the predictions of the (four) fine-tuned models (English, Chinese, Turkish, Russian): we average the class probability distributions of the models obtained for a target-language test instance, previously translated to each of the high-resource languages, respectively. We denote this \ttt ensemble approach with \rndhens.    
Since model ensembles are known to outperform single models \cite{wortsman2022model}, in our experiments, we compare \rndhens against the ensemble (of equally many models) fine-tuned on the round-trip translated source-language data (i.e., round-trip translated English) only, using different random seeds (we denote this with \rndsens). 

\subsection{Unsupported Languages}
\label{subsec:unsupported_languages}
Even though recent multilingual MT models cover a broad range of low-resource languages, the majority of the world's languages remain unsupported. Motivated by prior work on finding the best transfer source for a given target language \cite{lin-etal-2019-choosing,adelani-etal-2022-masakhaner,glavavs2021climbing}, we propose to translate to (\ttr) and from (\tts) an MT-supported language that is linguistically closest to the unsupported target: to this end, we quantify the linguistic proximity of languages as the cosine similarity of their typological vectors from the URIEL database \cite{littell-etal-2017-uriel}.
\section{Experimental Setup}
\label{sec:experimental_setup}

\rparagraph{Machine Translation} For translation, we leverage the state-of-the-art massively multilingual NLLB model with 3.3B parameters \cite{nllbteam2022language}. Building on prior work \cite{artetxe2023revisiting}, we ablate over decoding strategies, including greedy decoding, nucleus sampling with top-p = $0.8$, and beam search with beam size $5$. In our final evaluation, translations are generated using beam search. 

\rparagraph{Evaluation Tasks and Datasets.} Following prior work on low-resource XLT \cite{ansell-etal-2021-mad-g,ansell-etal-2022-composable,schmidt-etal-2022-dont}, we evaluate on sequence- and token-level classification tasks covering languages un(der)represented in the pretraining corpus of our base models. In all experiments, English is the source language.\footnote{We provide the complete list of languages in App.~ \ref{sec:models_and_datasets}.}

\rrparagraph{Natural Language Inference (NLI)} We evaluate our approaches on AmericasNLI (AmNLI) \cite{ebrahimi-etal-2022-americasnli}. AmNLI contains 10 indigenous languages of the Americas, only 3 of which are supported by the NLLB model we use \minute{and none are present in the pretraining corpus of our backbone model}. We utilize the English training and validation portion of XNLI \cite{conneau-etal-2018-xnli} as our source-language data. The dataset covers 393k training and 2490 validation instances. We jointly encode the hypothesis-premise pair and feed the transformed sequence start token into a feed-forward softmax classifier.

\rrparagraph{Text Classification (TC)} We use the sentiment classification dataset NusaX \cite{winata-etal-2023-nusax}, which comprises 10 languages from Indonesia, 7 of which are supported by the NLLB model \minute{and 2 are seen by our backbone model in pretraining}. The English training (500 instances) and validation portions (100 instances) are used as our source-language data. Similar to NLI, we feed the transformed representation of the sequence start token---output of the Transformer encoder---into the softmax classifier.

\rrparagraph{Named Entity Recognition (NER)} Our evaluation spans a set of 20 languages from MasakhaNER 2.0 (Masakha) \cite{adelani-etal-2022-masakhaner}. The dataset comprises a diverse set of underrepresented languages spoken in Sub-Saharan Africa. Among these, 18 languages are supported by the NLLB model we use for MT\minute{, but only 3 are covered in the pretraining corpus of our backbone model}. Our source data are the English training and validation portions of CoNLL \cite{tjong-kim-sang-de-meulder-2003-introduction}, with more than 14k instances for training and 3250 validation instances. In this token-level task, the classifier makes a prediction from the output (i.e., transformed) representation of each input token.

\rparagraph{Word Aligner}
Translation-based transfer for token-level tasks requires \textit{label projection}, i.e., mapping of the labels from source-language 
tokens to the tokens of the translated target sequence. To that end, we map labels post-translation with AccAlign \cite{wang-etal-2022-multilingual}, a state-of-the-art word aligner based on the multilingual sentence encoder LaBSE \cite{feng-etal-2022-language}.\footnote{We adhere to the hyperparameters specified in their work.} When recovering the labels for the translated sequences, we discard a training instance whenever we cannot map a labeled
source-language token to its target-language counterpart. The projection rates (for training data), i.e., the percentage of successful token mappings, is given for all supported languages in the App.~\ref{sec:proj}.

\rparagraph{Downstream Fine-Tuning}
We use XLM-R (\texttt{Large}) \cite{conneau-etal-2020-unsupervised} in all our experiments. For \tts and \ttt, we also experiment with RoBERTa (\texttt{Large}) \cite{liu2019roberta}. 
We outline the downstream fine-tuning details in Appendix \ref{sec:training_and_computation}. We evaluate models at various checkpoints of training: (i) at the end of the epoch\footnote{For AmNLI, we checkpoint after every 10\% of an epoch.} with the best performance on source-language validation data (\src), (ii) at the end of the epoch with the best performance on source-language validation data machine translated to the target language (\mtsrc), and (iii) at the end of the epoch with the best performance on target-language validation data (\oracle). \mtsrc and \oracle cannot be directly applied to \tts and \ttt as both model selection methods use (translated) target language data, while the training data of \tts and \ttt is solely in English. Hence, we adapt \mtsrc and \oracle for \tts and \ttt: for \mtsrc, we conduct round-trip translation on the source validation data pivoting through the target language (i.e., Source$\rightarrow$Target$\rightarrow$Source), and for \oracle, we simply MT-ed the (oracle) target validation data to the source language. Unless specified otherwise, we report results based on \src and show the results for \mtsrc and \oracle in Appendix \ref{sec:detailed_results}. We run experiments with 3 distinct random seeds and report mean accuracy for NLI and average F1 for NER and TC, as well as the standard deviation.
\section{Main Results and Discussion}

Table \ref{tab:train_vs_test_vs_traintest} summarizes our main results: performance of MT-based \ttr, \tts, and \ttt variants in low-resource XLT on three low-resource XLT benchmarks. 

\begin{table}[t!]
\small
\setlength{\tabcolsep}{2.3pt}
\centering
\begin{tabular}{llcccc}
\toprule
 &  & AmNLI & NusaX & Masakha & Avg  \\ 
 \midrule
\multicolumn{6}{c}{\textit{Zero-Shot}}               \\ \midrule
\source & X   & $44.7_{\pm1.2}$ & $71.2_{\pm1.3}$  & $47.9_{\pm0.6}$ & $54.6_{\pm1.1}$  \\ \midrule
\multicolumn{6}{c}{\textbf{\textit{Translate-Train}}}               \\ \midrule
\target             & X   & $61.1_{\pm0.4}$  & $77.8_{\pm0.8}$   & $62.1_{\pm0.3}$    & $67.0_{\pm0.5}$ \\
\jointt         & X   & $62.4_{\pm0.3}$  & $79.7_{\pm0.6}$ &  $\textbf{64.1}_{\pm0.3}$    & $\textbf{68.8}_{\pm0.4}$ \\
\tm       & X   & $63.4_{\pm0.5}$ & $79.0_{\pm0.7}$ & $56.9_{\pm0.4}$ & $66.4_{\pm0.5}$ \\
\jointtm   & X   & $63.6_{\pm0.6}$ & $80.8_{\pm0.4}$ & $57.4_{\pm0.6}$ & $67.3_{\pm0.5}$ \\ 
\midrule
\multicolumn{6}{c}{\textit{incl. Translations to High-Resource Languages}}           \\
\midrule
\jointth             & X   & $62.9_{\pm0.5}$  & $78.1_{\pm1.3}$  & $62.9_{\pm0.3}$ & $68.0_{\pm0.8}$  \\
\jointtmh        & X   & $64.7_{\pm0.4}$ & $79.1_{\pm1.9}$ & $58.0_{\pm0.5}$ & $67.3_{\pm1.2}$  \\ \midrule
\multicolumn{6}{c}{\textit{\textbf{Translate-Test}}}                \\ \midrule
\source             & R   & $53.1_{\pm0.1}$  & $79.4_{\pm0.4}$  & $54.7_{\pm0.1}$    & $62.4_{\pm0.2}$ \\
\source             & X & $52.9_{\pm0.5}$  & $80.9_{\pm0.8}$  & $54.1_{\pm0.1}$    & $62.6_{\pm0.5}$ \\ \midrule
\multicolumn{6}{c}{\textit{\textbf{Roundtrip-Train-Test}}}          \\ \midrule
\rndt        & R   & $62.4_{\pm0.6}$  & $81.2_{\pm0.4}$  & $54.6_{\pm0.1}$    & $66.1_{\pm0.4}$ \\
\rndt        & X & $63.1_{\pm0.4}$  & $81.6_{\pm0.5}$  & $53.6_{\pm0.2}$    & $66.1_{\pm0.4}$ \\
\rndtm  & R   & $64.3_{\pm0.2}$  & $81.0_{\pm0.4}$  & $54.0_{\pm0.2}$    & $66.4_{\pm0.3}$ \\
\rndtm  & X & $64.0_{\pm0.3}$  & $82.1_{\pm0.4}$  & $53.0_{\pm0.4}$ & $66.4_{\pm0.4}$ \\ 
\rndsens & X   & $63.7_{\pm0.2}$  & $82.8_{\pm0.3}$  & $53.7_{\pm0.1}$  & $66.7_{\pm0.2}$   \\ 
\midrule
\multicolumn{6}{c}{\textit{incl. Translations to High-Resource Languages}}           \\
\midrule
\rndhens & X   & $\textbf{66.1}_{\pm0.2}$  & $\textbf{83.9}_{\pm0.4}$  & $45.8_{\pm0.1}$ &  $65.3_{\pm0.3}$   \\
\bottomrule
\end{tabular}
\caption{Results for translation-based XLT for languages supported by the MT model. We use XLM-R (X) and RoBERTa (R). The best results are shown in \textbf{bold}. 
}
\label{tab:train_vs_test_vs_traintest}
\vspace{-1em}
\end{table}
\sparagraph{\ttr vs. \tts}
We first assess the widely used \ttr and \tts baselines. These simple translation-based XLT strategies outperform ZS-XLT dramatically: from 6.2\% on Masakha (\tts with XLM-R) up to 18.9\% on AmNLI (\jointtm), rendering them as unavoidable baselines for any XLT effort. Keeping the original clean source language data in the training mix is beneficial: \jointt and \jointtm consistently outperform \target and \tm, respectively.
For sequence-level classification tasks (AmNLI and NusaX), training on the concatenation of the clean source data and the source data translated to a set of related target languages (\jointtm) yields the best results. For NER on Masakha, \jointt maximizes XLT performance.
We further observe that the optimal \ttr (\jointt) strategy significantly outperforms (+6.2\%) the best \tts approach.
Our \tts results also demonstrate that in low-resource XLT, mLMs yield comparable performance to monolingual LMs: this contradicts the recent \tts finding for high-resource languages of \citet{artetxe2023revisiting}. 

\rparagraph{\ttt} For sequence-level classification tasks, we find that \ttt outperforms the best \ttr strategy (\jointtm), which is in line with prior findings \cite{artetxe2023revisiting, oh-etal-2022-synergy}. For NusaX, this observation holds for all \ttt variants. For AmNLI, only \rndtm consistently outperforms \jointtm.
We further observe inconclusive results regarding the LM for which we get the highest performance for \rndtm: while RoBERTa is superior on AmNLI, XLM-R displays better performance on NusaX.
This result, however, does not extend to \rndt, for which XLM-R consistently outperforms RoBERTa.
As already seen, \tts lags \ttr on Masakha, and this is also true for \ttt. Even more so, \ttt progressively degrades in performance the more round-trip translated data is introduced (i.e., \rndt trails \tts by at least 0.1\% whereas \rndtm does so by 0.7\%).
We hypothesize that both the amount of round-trip translated data and the type of task drive the performance of monolingual LMs like RoBERTa in translation-based XLT to low-resource languages. Our results challenge prior work \cite{artetxe2023revisiting, oh-etal-2022-synergy}, in which \tts and \ttt are better with monolingual LMs than with mLMs. Their experiments, however, covered predominantly high-resource target languages.

\rparagraph{Adding High-Resource Languages}
Table \ref{tab:train_vs_test_vs_traintest} further reports results of \ttr and \ttt variants that include high-resource languages (i.e., Chinese, Russian, and Turkish) for translation-based XLT.
The results for \ttr are inconsistent.
For AmNLI, including high-resource languages (\jointtmh) boosts performance by at least 1.1\%. These gains persist for different model selection strategies (cf. Appendix \ref{sec:detailed_results}).
However, such multilingual data augmentation adversely affects the performance on NusaX and Masakha.
We posit that the choice of high-resource languages critically affects \ttr since the test data is still in the low-resource target language, increasing the risk of negative transfer.
In contrast, integrating high-resource languages into \ttt (i.e., \rndhens) results in substantial gains of at least 1.8\% for AmNLI and NusaX compared to \rndtm.
Unlike its success on sequence-level classification tasks, \rndhens degrades performance for Masakha.
While ensembles often inherently produce higher scores than single models \cite{wortsman2022model}, our results on sequence-level tasks show that ensembles trained on round-trip translations to various high-resource languages (\rndhens) outperform ensembles trained solely on round-trip translated data to the source language (\rndsens). In contrast to \ttr, integrating high-resource languages in \ttt reduces the likelihood of negative transfer since the test data is in the same language as the training data.
Ensembling additionally smooths over language-specific translation and downstream transfer errors. Finally, ensembling monolingual LMs might offer further gains but requires such models for each high-resource language.

\rparagraph{MT-Strategies for Model Selection} 
\begin{table}[t!]
\small
\setlength{\tabcolsep}{5.8pt}
\centering
\begin{tabular}{@{}lcccc@{}}
\toprule
 & AmNLI & NusaX & Masakha & Avg  \\ \midrule
\multicolumn{5}{c}{\textbf{\textit{Translate-Train}}} \\ \midrule
\srct              & $62.6_{\pm0.5}$  & $79.3_{\pm0.6}$  & $60.1_{\pm0.4}$    & $67.3_{\pm0.5}$ \\
\mtsrct            & $62.8_{\pm0.5}$  & $79.6_{\pm0.7}$  & $60.3_{\pm0.3}$    &  $67.6_{\pm0.5}$ \\
\oraclet           & $\underline{62.9}_{\pm0.5}$  & $\underline{80.2}_{\pm0.6}$  & $\underline{62.2}_{\pm0.4}$    & $\textbf{68.4}_{\pm0.5}$ \\ \midrule
\multicolumn{5}{c}{\textbf{\textit{Translate-Test}}} \\ \midrule
\srct              & $53.0_{\pm0.4}$  & $80.1_{\pm0.6}$  &  $\underline{54.4}_{\pm0.6}$   & $62.5_{\pm0.5}$ \\
\mtsrct            & $53.1_{\pm0.5}$  & $79.8_{\pm0.5}$  & $54.3_{\pm0.1}$   & $62.4_{\pm0.4}$ \\
\oraclet           & $\underline{53.4}_{\pm0.4}$  & $\underline{80.8}_{\pm0.7}$   & $\underline{54.4}_{\pm0.1}$    & $\textbf{62.9}_{\pm0.5}$ \\ \midrule
\multicolumn{5}{c}{\textbf{\textit{Roundtrip-Train-Test}}} \\ \midrule
\srct              & $\underline{63.5}_{\pm0.4}$  & $81.5_{\pm0.4}$  & $53.8_{\pm0.3}$    & $66.3_{\pm0.4}$ \\
\mtsrct            & $\underline{63.5}_{\pm0.4}$  & $81.4_{\pm0.5}$  & $53.7_{\pm0.2}$    & $66.2_{\pm0.4}$\\
\oraclet           & $63.4_{\pm0.5}$  & $\underline{81.7}_{\pm0.4}$  & $\underline{54.0}_{\pm0.2}$    & $\textbf{66.4}_{\pm0.4}$ \\
\bottomrule
\end{tabular}
\caption{Comparison of model selection strategies for languages supported by the MT model. We average the results of \target, \jointt, \tm, and \jointtm for \ttr, \source for \tts, and \rndt and \rndtm for \ttt. The best results per task and training setup (e.g., \ttr) are \underline{underlined}; the best results for each training setup are shown in \textbf{bold}.}
\label{tab:ckpt}
\end{table}
In XLT, model selection is done using validation data in the source or target language, with the latter violating true ZS-XLT \cite{schmidt-etal-2022-dont,schmidt-etal-2023-free}. The usage of MT to create validation data for model selection, however, remains understudied \cite{ebrahimi-etal-2022-americasnli}. We thus next explore MT-based model selection strategies and compare them against standard counterparts (cf. \S\ref{sec:experimental_setup}) in Table \ref{tab:ckpt}. In \ttr, in line with prior work \cite{ebrahimi-etal-2022-americasnli,schmidt-etal-2022-dont}, \oracle outperforms all other model selection variants. We show, however, for the first time, that it is also the upper bound of \tts and \ttt.  
Additionally, in \ttr \mtsrc (i.e., model selection based on the automatically translated target language validation data) surpasses \src on average across all tasks; this is notably not the case for \tts and \ttt.  

\rparagraph{Unsupported Languages}
\begin{table}[t!]
\small
\setlength{\tabcolsep}{3.3pt}
\centering
\begin{tabular}{@{}llcccc@{}}
\toprule
\multicolumn{1}{c}{}   &         & AmNLI & NusaX & Masakha & Avg  \\ \midrule
\multicolumn{6}{c}{\textit{Zero-Shot}}               \\ \midrule
\source                    & X   & $44.2_{\pm0.6}$  & $57.8_{\pm1.4}$  & $60.2_{\pm1.6}$    & $54.1_{\pm1.3}$ \\ \midrule
\multicolumn{6}{c}{\textbf{\textit{Translate-Train}}}                      \\ \midrule
\jointt                & X   & $\textbf{47.5}_{\pm0.4}$  & $67.5_{\pm1.3}$  &  $\textbf{61.8}_{\pm0.8}$    & $59.0_{\pm0.9}$ \\
\jointtm          & X   & $46.5_{\pm0.3}$  & $\textbf{74.0}_{\pm1.4}$  & $61.0_{\pm1.2}$    & $\textbf{60.5}_{\pm1.1}$ \\ \midrule
\multicolumn{6}{c}{\textbf{\textit{Translate-Test}}}                       \\ \midrule
\source                    & R & $36.5_{\pm0.2}$  & $54.4_{\pm1.3}$  & $48.1_{\pm0.5}$    & $46.3_{\pm0.8}$ \\
\source                    & X   & $37.4_{\pm0.3}$  & $54.9_{\pm1.5}$  & $46.6_{\pm1.4}$    & $46.3_{\pm1.2}$ \\ \midrule
\multicolumn{6}{c}{\textbf{\textit{Roundtrip-Train-Test}}}                 \\ \midrule
\rndtm         & R & $38.8_{\pm0.3}$  & $60.5_{\pm0.7}$  & $45.0_{\pm0.4}$    & $48.1_{\pm0.5}$ \\
\rndtm         & X   & $39.1_{\pm0.2}$  & $59.1_{\pm1.2}$  & $44.0_{\pm1.4}$    & $47.4_{\pm1.1}$ \\
\midrule
\multicolumn{6}{c}{\textit{incl. Translations to High-Resource Languages}}           \\
\midrule
\rndhens & X   & $41.1_{\pm0.2}$  & $65.0_{\pm0.6}$  & $42.8_{\pm0.6}$ &  $49.6_{\pm0.5}$  \\ \bottomrule
\end{tabular}
\caption{Results for translation-based XLT for languages \textbf{not} supported by the MT model. We use XLM-R (X) and RoBERTa (R). The best results are shown in \textbf{bold}. 
}
\label{tab:unsupported}
\end{table}
Even the most multilingual MT models \cite{nllbteam2022language} support only a tiny fraction of the world's 7000+ languages. Table \ref{tab:unsupported} summarizes the performance of our MT-based XLT strategy for languages not supported by MT, where we translate to/from the closest respective supported language (see \S\ref{subsec:unsupported_languages}).  
We find that \ttr strategies remain successful and substantially improve by 4.9\% (\jointt) and 6.4\% (\jointtm) over the ZS-XLT on average.
In contrast, \tts and \ttt for unsupported languages substantially trail ZS-XLT performance. This is because it is not really possible to get good translations in the source language by simply pretending the input comes from a different, supported language (in \tts and \ttt). In contrast, with \ttr, we obtain proper translations in a supported language that is close to the real target (as in \ttr): the transfer then amounts to the mLM-based ZS-XLT ability from the close, MT-supported language to the real MT-unsupported target.   
This further supports the finding that MT quality much less affects performance of \ttr strategies than of \tts or \ttt approaches \cite{artetxe2023revisiting}. 
\section{Further Findings}

\sparagraph{Decoding Strategy}
\begin{table}[t!]
\small
\setlength{\tabcolsep}{7.5pt}
\centering
\begin{tabular}{@{}lcccc@{}}
\toprule
        & AmNLI & NusaX & Masakha & Avg           \\ \midrule
Nucleus & $56.2_{\pm3.2}$  & $75.6_{\pm2.4}$  & $60.5_{\pm1.9}$    &  $64.1_{\pm2.6}$         \\
Greedy  & $62.5_{\pm0.6}$  & $\textbf{79.5}_{\pm2.2}$  & $ 64.0_{\pm1.1}$   & $68.7_{\pm1.5}$\\
Beam    & $\textbf{62.6}_{\pm0.5}$  & $79.4_{\pm2.1}$  & $\textbf{64.8}_{\pm1.2}$    &  $68.9_{\pm1.4}$\\ \bottomrule
\end{tabular}
\caption{Results for \ttr (\target) for different decoding strategies evaluated on the validation data of AmNLI, NusaX, and Masakha. The best results are shown in \textbf{bold}.}
\label{tab:translation}
\end{table}
Previous work examined the impact of various decoding strategies on downstream performance, particularly in the context of back-translation \cite{edunov-etal-2018-understanding} and sequence-level classification \cite{artetxe2023revisiting}. They found nucleus sampling consistently superior to beam search and greedy decoding. However, our results in Table \ref{tab:translation} suggest a noteworthy deviation for low-resource languages. We find beam search and greedy decoding substantially outperform nucleus sampling. We posit that the underrepresentation of low-resource languages in the training data of MT models contributes to this contrast.\footnote{We present details on the resource availability of the tasks we evaluated, compared to related work, in Appendix \ref{ref:res}.} 

\rparagraph{Joint vs. Sequential Training} 
\begin{table}[t!]
\small
\setlength{\tabcolsep}{6.4pt}
\centering
\begin{tabular}{@{}lcccc@{}}
\toprule
 & AmNLI & NusaX & Masakha & Avg  \\ \midrule
Joint                & $63.5_{\pm0.4}$  &  $\textbf{80.7}_{\pm0.4}$ & $\textbf{62.8}_{\pm0.4}$    & $\textbf{69.0}_{\pm0.4}$ \\
Sequential           & $\textbf{64.1}_{\pm1.4}$ &  $80.1_{\pm0.7}$ & $62.4_{\pm0.4}$    & $68.9_{\pm0.9}$ \\ \bottomrule
\end{tabular}
\caption{Comparison of sequential vs. joint translation-based XLT for languages supported by the MT model. We average the results of \jointt and \jointtm and the respective sequential variants (\seqt and \seqtm). The best results are shown in \textbf{bold}. Model selection is done on the best epoch based on target language validation data (\oracle). 
}
\label{tab:joint_vs_seq}
\end{table}
Prior work primarily concatenated the source data with the translated target language data and trained on both jointly \cite{pmlr-v119-hu20b,oh-etal-2022-synergy,artetxe2023revisiting}. In contrast, \citet{aggarwal-etal-2022-indicxnli} propose a sequential \ttr approach in which the model is first trained on the source-language data and then, in a subsequent step, on the translated data of either (i) a single target language (\target) or (ii) multiple target languages jointly (\tm). We adopt this in our \ttr variants (denoted \seqt and \seqtm) and compare them against the more established joint training: results in Table \ref{tab:joint_vs_seq} show comparable performance between the two. This favors sequential training, as it is more computationally efficient \cite{schmidt-etal-2022-dont}.  

\rparagraph{Importance of the MT model} 
\begin{table}[]
\small
\setlength{\tabcolsep}{1.3pt}
\begin{tabular}{@{}llcccccccc@{}}
\toprule
 & &\multicolumn{2}{c}{AmNLI} & \multicolumn{2}{c}{NusaX} & \multicolumn{2}{c}{Masakha} & \multicolumn{2}{c}{Avg} \\ \cmidrule{3-10}
 & &NLLB        & GT          & NLLB        & GT          & NLLB         & GT           & NLLB       & GT         \\ \midrule
\multicolumn{10}{c}{\textbf{\textit{Translate-Train}}}                                                                     \\ \midrule
\jointt & X & 62.9        & 63.1        & 87.0        & 87.0        & 66.0         & 65.3         & 72.0       & 71.8       \\ \midrule
\multicolumn{10}{c}{\textbf{\textit{Translate-Test}}}                                                                      \\ \midrule
\source & R & 53.1        & 63.6        & 85.3        & 85.7        & 54.8         & 59.7         &     64.4       &     69.7       \\
\source & X & 52.9        & 64.8        & 85.8        & 86.6        & 54.1         & 59.0         &     64.3       &     70.1       \\ \midrule
\multicolumn{10}{c}{\textbf{\textit{Roundtrip-Train-Test}}}                                                                \\ \midrule
\rndt & R & 62.4        &     69.9        & 85.2        & 86.9        & 55.0         & 59.9         &    67.5        &   72.2         \\
\rndt & X & 63.1        &     69.1        & 86.0        & 87.5        & 54.0         & 59.2         &    67.7        &   71.9          \\ \bottomrule
\end{tabular}
\caption{Results for translation-based XLT for two MT systems (NLLB and GT) for languages supported by both MT models. For \ttr, model selection is done on the best epoch based on target-language validation data (\oracle), and for \tts and \ttt, based on source-language validation data (\src). We evaluated XLM-R (X) and RoBERTa (R).}
\label{tab:google}
\end{table}
\minute{In the MT landscape, the translation quality of the industrial-grade models is often considered superior to that of their publicly available counterparts. Because of this, we ablate the impact of the used MT model on translation-based XLT by generating translations through Google Translate (GT)---a representative example of an industrial MT model. We evaluate \ttr (i.e., \jointt), \tts (i.e., \source), and \ttt (i.e., \rndt) with GT. Our results in Table \ref{tab:google} indicate that the performance remains comparable for \ttr, while GT surpasses NLLB by a substantial margin in the context of \tts and \ttt. We hypothesize that our observation stems from the increased translation quality which is of larger importance for \tts and \ttt than for \ttr \cite{artetxe2023revisiting}. Furthermore, our ablation confirms that \ttt remains the most competitive translation-based XLT method for sequence-level classification tasks. Unfortunately, GT does not support the exact same set of languages as NLLB. We thus carry out the ablation on the following languages supported by both MT systems: for AmNLI: Aymara, Guarani, and Quechua; for NusaX: Javanese and Sundanese; and for Masakha: Bambara, Éwé, Hausa, Igbo, Kinyarwanda, chiShona, Kiswahili, Akan/Twi, isiXhosa, Yorùbá, isiZulu.}
\section{Related Work}
\sparagraph{Translation-based Transfer}
Translation-based XLT has been adopted early \cite{fortuna2005use, banea-etal-2008-multilingual, shi-etal-2010-cross} yet remains a competitive baseline to date \cite{ruder-etal-2021-xtreme,ebrahimi-etal-2022-americasnli,aggarwal-etal-2022-indicxnli}. 
Prior work evaluated training on the translated data of a single target language \cite{ebrahimi-etal-2022-americasnli}, on the concatenation of all target languages \cite{ruder-etal-2021-xtreme}, and have integrated the source language either by sequentially training first on the source followed by the translated target language data \cite{aggarwal-etal-2022-indicxnli} or by jointly training on the concatenation of both \cite{chen-etal-2023-frustratingly}. While earlier approaches focus primarily on the translation of the training data (\ttr), more recent work evaluated the translation of test data as well \cite{pmlr-v119-hu20b, isbister-etal-2021-stop} (\tts). Finally, both approaches can be combined by training the model on round-trip translated noisy source data (i.e., translating source data to the target language and back to the source) and evaluating it on target language test data translated to the source language \cite{artetxe-etal-2020-translation, oh-etal-2022-synergy,artetxe2023revisiting}. Previous studies have either focused on improving one of these paradigms or utilized them as baselines. In contrast, we provide a comparative empirical evaluation of existing translation-based approaches to XLT, testing them explicitly against ZS-XLT for low-resource languages.

\rparagraph{Label projection} Translation-based transfer for token-level tasks necessitates label projection, which is achieved through either alignment-based or \cite{tjong-kim-sang-de-meulder-2003-introduction, jalili-sabet-etal-2020-simalign, nagata-etal-2020-supervised} marker-based approaches \cite{lee-etal-2018-semi, lewis-etal-2020-mlqa, pmlr-v119-hu20b, bornea2021multilingual}. The former maps each token in the source sequence to a token in the translated target sequence, with recent neural word aligners utilizing contextualized embeddings of mLMs to produce the alignment \cite{dou-neubig-2021-word, wang-etal-2022-multilingual}. Marker-based alignment, in contrast, entails marking labeled tokens in the sequence prior to translation, often by enclosing them in XML or HTML tags, and preserving them throughout the translation process. Subsequently, the labels can be recovered from the markers. While alignment-based methods are prone to issues like error propagation, translation shift \cite{akbik-etal-2015-generating}, and non-contiguous alignments \cite{zenkel-etal-2020-end}, marker-based projection compromises translation performance by introducing artificial tokens and is susceptible to vanishing markers, particularly with non-industrial, publicly available translation models \cite{chen-etal-2023-frustratingly}. 
In XLT for NER (Masakha), we leveraged a state-of-the-art alignment-based model \cite{wang-etal-2022-multilingual}. 

\section{Conclusion}
We reviewed the field of translation-based cross-lingual transfer (XLT) to low-resource languages through a comparative evaluation of various approaches---derived from translate-train (\ttr), translate-test (\tts), and roundtrip-train-test (\ttt)---on three established benchmarks encompassing 40 languages.
We demonstrated that translation-based XLT substantially outperforms zero-shot XLT no matter the task. Furthermore, irrespective of the translation-based strategy, including the clean source language data in the training yielded consistent improvements. 
For sequence-level tasks, training on the source language data round-trip translated through a set of related target languages and evaluating, at inference, the target language instances translated back to the source language performed best (\ttt). In contrast, for token-level tasks, training on the translations to a single target language showed the best results (\ttr). Additionally, we proposed novel translation-based XLT strategies for \ttr and \ttt by including translations to a set of typologically diverse high-resource languages. 
Further, we successfully proposed translation-based strategies for languages unsupported by the MT model and showcased the effectiveness of using automatically translated validation data for model selection. Our empirical comparison and its findings warrant broader inclusion of more competitive translation-based XLT approaches as standard baselines in all research efforts set to improve XLT with mLMs.

\section{Limitations}
We strove to provide a comprehensive and systematic evaluation of translation-based XLT to low-resource languages, additionally providing novel \ttr and \ttt paradigms. However, our study faces limitations, primarily stemming from the prevalent practice of obtaining benchmarks for low-resource languages by translating datasets from high-resource languages, which applies to AmNLI, NusaX, and some languages of Masakha. The resulting data possesses distinctive characteristics arising from the translation process, commonly referred to as \textit{translationese}. On the one hand, we explicitly exploit this behavior by demonstrating that augmenting the training data in the same way as we augment the test data (i.e., \ttt) yields the best results. On the other hand, there exist uncontrollable implications potentially influencing our results, for instance, that translation often becomes easier for datasets originating from translation themselves.

\section*{Acknowledgements}

This work was in part supported by the Alcatel-Lucent Stiftung (Grant ``EQUIFAIR: Equitably Fair and Trustworthy Language Technology'').   

\bibliography{anthology,custom}

\appendix

\newpage
\section{Models and Datasets}
\label{sec:models_and_datasets}
\begin{table}[h!]
\small
\centering
\setlength{\tabcolsep}{2.8pt}
\begin{tabular}{@{}llccc@{}}
\toprule
Language       & Code & \multicolumn{1}{l}{Pre. Model} & \multicolumn{1}{l}{Supp. Trans.} & \multicolumn{1}{l}{Closest} \\ \midrule
\multicolumn{5}{c}{\textbf{AmNLI}}                                                                                                         \\
\midrule
Aymara         & AYM  & No                                    & Yes                                  & -                                    \\
Guaraní        & GN   & No                                    & Yes                                  & -                                    \\
Quechua        & QUY  & No                                    & Yes                                  & -                                    \\
Asháninka      & CNI  & No                                    & No                                 & AYM                                  \\
Bribri         & BZD  & No                                    & No                                 & QUY                                  \\
Nahuatl        & NAH  & No                                    & No                                 & GN                                   \\
Otomí          & OTO  & No                                    & No                                 & GN                                   \\
Rarámuri       & TAR  & No                                    & No                                 & AYM                                  \\
Shipibo-Konibo & SHP  & No                                    & No                                 & QUY                                  \\
Wixarika       & HCH  & No                                    & No                                 & GN                                   \\
\midrule
\multicolumn{5}{c}{\textbf{NusaX}}                                                                                                        \\
\midrule
Acehnese       & ACE  & No                                    & Yes                                  & -                                    \\
Balinese       & BAN  & No                                    & Yes                                  & -                                    \\
Banjarese      & BJN  & No                                    & Yes                                  & -                                    \\
Buginese       & BUG  & No                                    & Yes                                  & -                                    \\
Minangkabau    & MIN  & No                                    & Yes                                  & -                                    \\
Javanese       & JAV  & Yes                                     & Yes                                  & -                                    \\
Sundanese      & SUN  & Yes                                     & Yes                                  & -                                    \\
Madurese       & MAD  & No                                    & No                                 & SUN                                  \\
Ngaju          & NIJ  & No                                    & No                                 & SUN                                  \\
Toba Batak     & BBC  & No                                    & No                                 & BUG                                  \\
\midrule
\multicolumn{5}{c}{\textbf{MasakhaNER}}                                                                                                   \\
\midrule
Bambara        & BAM  & No                                    & Yes                                  & -                                    \\
Éwé            & EWE  & No                                    & Yes                                  & -                                    \\
Fon            & FON  & No                                    & Yes                                  & -                                    \\
Hausa          & HAU  & Yes                                     & Yes                                  & -                                    \\
Igbo           & IBO  & No                                    & Yes                                  & -                                    \\
Kinyarwanda    & KIN  & No                                    & Yes                                  & -                                    \\
Luganda        & LUG  & No                                    & Yes                                  & -                                    \\
Luo            & LUO  & No                                    & Yes                                  & -                                    \\
Mossi          & MOS  & No                                    & Yes                                  & -                                    \\
Chichewa       & NYA  & No                                    & Yes                                  & -                                    \\
chiShona       & SNA  & No                                    & Yes                                  & -                                    \\
Kiswahili      & SWA  & Yes                                     & Yes                                  & -                                    \\
Setswana       & TSN  & No                                    & Yes                                  & -                                    \\
Akan/Twi       & TWI  & No                                    & Yes                                  & -                                    \\
Wolof          & WOL  & No                                    & Yes                                  & -                                    \\
isiXhosa       & XHO  & Yes                                     & Yes                                  & -                                    \\
Yorùbá         & YOR  & No                                    & Yes                                  & -                                    \\
isiZulu        & ZUL  & No                                    & Yes                                  & -                                    \\
Ghomálá’       & BBJ  & No                                    & No                                 & SWA                                  \\
Naija          & PCM  & No                                    & No                                 & HAU                                  \\ \bottomrule
\end{tabular}
\caption{List of languages per task showing the coverage in the pretraining corpus of our backbone model (\textit{Pre. Model}), the support by NLLB (\textit{Supp. Trans.}), and the closest language we translated to/from for languages that are not supported by the translation model (\textit{Closest}).}
\label{tab:lang_list}
\end{table}

The models for translation, word alignment, and downstream fine-tuning were accessed through the Hugging Face transformers library \cite{wolf-etal-2020-transformers}. Additional adapter checkpoints for the used word aligner were downloaded from the corresponding GitHub repository: \href{https://github.com/sufenlp/AccAlign}{AccAlign} \cite{wang-etal-2022-multilingual}. We accessed all our datasets through the Hugging Face datasets library \cite{lhoest-etal-2021-datasets}. Further, we ensured compliance with the licenses of the models and datasets. Table \ref{tab:lang_list} displays a detailed list of all languages.




\section{Word Alignment}
\label{sec:proj}
Table \ref{tab:proj} shows the projection rates for AccAlign \cite{wang-etal-2022-multilingual} (used in our work) and the state-of-the-art marker-based method EasyProject (EasyProj) \cite{chen-etal-2023-frustratingly}. The projection rate is computed as the ratio of retained training instances after label projection to all instances in the original training data. The results highlight that the downstream performance of AccAlign is on par with the competitive EasyProj. 
Nevertheless, we attribute variations in the projection rate not only to superior alignment but also to differences in filtering strategies. While \citet{chen-etal-2023-frustratingly} filter translated instances that do not match the number and type of tags in the source instance, our approach filters instances if a tagged source-language token cannot be mapped to its target language equivalent. We leave the exploration of the impact of different filtering approaches to future work.

\begin{table}[t!]
\small
\setlength{\tabcolsep}{22pt}
\centering
\begin{tabular}{@{}lcc@{}}
\toprule
                 & AccAlign & EasyProj \\ \midrule
BAM              & 94.4     & 90.9     \\
EWE              & 95.6     & 92.2     \\
FON              & 92.9     & 83.4     \\
HAU              & 97.5     & 94.4     \\
IBO              & 98.3     & 96.1     \\
KIN              & 97.2     & 93.8     \\
LUG              & 97.0     & 95.3     \\
LUO              & 96.6     & 94.0       \\
MOS              & 90.3     & 92.3     \\
NYA              & 98.5     & 96.7     \\
SNA              & 98.7     & 95.6     \\
SWA              & 98.8     & 96.3     \\
TSN              & 98.0     & 95.0       \\
TWI              & 96.2     & 94.6     \\
WOL              & 93.0     & 93.4     \\
XHO              & 97.8     & 95.1     \\
YOR              & 97.3     & 94.3     \\
ZUL              & 97.7     & 93.1     \\ \midrule
Avg Proj. Rate   & 96.4     & 93.7     \\ \midrule
Avg. Perf.  & 65.5     & 65.6     \\ \bottomrule
\end{tabular}
\caption{Projection rates and average performance in the \jointt setup for word alignments produce by AccAlign \cite{wang-etal-2022-multilingual} and EasyProj \cite{chen-etal-2023-frustratingly}. Model selection is done on the best epoch based on target-language validation data (\oracle).}
\label{tab:proj}
\end{table}

\section{Training and Computational Details}
\label{sec:training_and_computation}
Table \ref{tab:hp} outlines the hyperparameters for downstream fine-tuning of our utilized tasks.\footnote{We used a comparably small learning rate for AmNLI as single seeds did not converge for higher learning rates in preliminary experiments.} Alongside, we implement a linear schedule of 10\% warm-up and decay and employ mixed precision. All translations were run on a single A100 with 40GB VRAM, and all downstream training and evaluation runs were completed on a single V100 with 32GB VRAM. We roughly estimate that GPU time accumulates to 3500 hours across all translations and downstream fine-tunings.

\begin{table}[t!]
\small
\setlength{\tabcolsep}{13.5pt}
\begin{tabular}{@{}lccc@{}}
\toprule
              & AmNLI & NusaX & Masakha \\ \midrule
Task          & NLI   & TC    & NER     \\
Epochs        & 2     & 20    & 10      \\
Batch Size    & 32    & 32    & 32      \\
Learning Rate & 2e-6  & 1e-5  & 1e-5    \\
Weight Decay  & 0.01  & 0.01  & 0.01    \\ \bottomrule
\end{tabular}
\caption{Hyperparameters for downstream fine-tuning.}
\label{tab:hp}
\end{table}

\section{Resource Availability}
\label{ref:res}
\begin{table*}[t!]
\small
\setlength{\tabcolsep}{4.1pt}
\begin{tabular}{@{}lccccccc|cccc@{}}
\toprule
                       & \multicolumn{7}{c|}{\citet{artetxe2023revisiting}}                                 & \multicolumn{4}{c}{Ours}       \\ \cmidrule{2-8} \cmidrule{9-12}
                       & XNLI & PAWS-X & MARC & XStory & XCOPA & EXAMS & Avg  & AmNLI & NusaX & Masakha & Avg  \\ \midrule
Avg. Res. Availability & 2.67 & 3.61   & 4.24 & 2.24   & 1.17  & 2.67  & 2.78 & 0.09  & 0.21  & 0.41    & 0.24 \\ \bottomrule
\end{tabular}
\caption{Average percentage of available parallel data per task from the corpus used to train NLLB for three datasets we evaluated on: AmNLI, NusaX, and Masakha; and six datasets \citet{artetxe2023revisiting} did: XNLI \cite{conneau-etal-2018-xnli}, PAWS-X \cite{yang-etal-2019-paws}, MARC \cite{keung-etal-2020-multilingual}, XStoryCloze (XStory) \cite{lin-etal-2022-shot}, XCOPA \cite{ponti-etal-2020-xcopa}, EXAMS \cite{hardalov-etal-2020-exams}.}
\label{tab:translation_data}
\end{table*}
To substantiate our claim that the languages we evaluate are characterized by far lower resource availability compared to related work, we assess the relative size of parallel data used for training NLLB for languages encompassed in the datasets we used and those employed in \citet{artetxe2023revisiting}. For each language, we calculate the ratio of available parallel data to the total size of the parallel corpus and, subsequently, average the results per dataset. The computations are based on the following corpus \href{https://huggingface.co/datasets/allenai/nllb}{https://huggingface.co/datasets/allenai/nllb}. Our metric serves as a proxy for the average coverage of a dataset in the training data of NLLB. As shown in Table \ref{tab:translation_data}, the resource availability for the datasets we evaluated is approximately an order of magnitude smaller.

\begin{table*}[t!]
\section{Detailed Main Results}
\label{sec:detailed_results}
\small
\setlength{\tabcolsep}{7.8pt}
\centering
\begin{tabular}{@{}llcccccccccccc@{}}
\toprule
                    &         & \multicolumn{3}{c}{AYM} & \multicolumn{3}{c}{GN} & \multicolumn{3}{c}{QUY} & \multicolumn{3}{c}{Avg} \\ \cmidrule{3-5} \cmidrule{6-8} \cmidrule{9-11} \cmidrule{12-14}
                    &         & I   & II  & III   & I   & II  & III  & I   & II  & III   & I   & II  & III   \\ \midrule
\multicolumn{14}{c}{\textit{Zero-Shot}}                                                                                              \\ \midrule
\source                   & X   & 43.2  & 44.0    & 42.4  & 46.5  & 46.8    & 47.7 & 44.3  & 44.7    & 44.2  & 44.7  & 45.2    & 44.8  \\ \midrule
\multicolumn{14}{c}{\textbf{\textit{Translate-Train}}}                                                                                        \\ \midrule
\jointh                & X   & 38.0  & 38.8    & 38.8  & 42.0  & 44.8    & 44.5 & 40.2  & 42.1    & 41.7  & 40.1  & 41.9    & 41.7  \\
\texttt{T}                    & X   & 58.4  & 59.5    & 58.7  & 63.6  & 63.2    & 62.8 & 61.5  & 62.2    & 61.8  & 61.1  & 61.6    & 61.1  \\
\jointt                & X   & 59.4  & 59.4    & 59.6  & 66.1  & 65.6    & 65.6 & 61.9  & 62.3    & 63.4  & 62.4  & 62.4    & 62.9  \\
\seqt  & X   & 53.8  & 62.8    & 61.9  & 64.1  & 66.2    & 67.0 & 54.8  & 64.3    & 62.7  & 57.6  & 64.4    & 63.9  \\
\jointth             & X   & 59.4  & 59.7    & 59.8  & 65.8  & 65.8    & 66.2 & 63.5  & 63.9    & 64.3  & 62.9  & 63.1    & 63.4  \\
\tm                & X   & 61.2  & 61.6    & 61.4  & 64.4  & 64.1    & 64.2 & 64.7  & 64.4    & 64.7  & 63.4  & 63.4    & 63.5  \\
\jointtm               & X   & 61.4  & 62.4    & 62.3  & 65.5  & 65.2    & 65.2 & 63.8  & 64.0    & 64.8  & 63.6  & 63.9    & 64.1  \\
\seqtm& X   & 58.3  & 62.1    & 62.6  & 60.6  & 66.8    & 66.8 & 59.5  & 65.0    & 65.0  & 59.5  & 64.7    & 64.8  \\
\jointtmh            & X   & 62.7  & 63.0    & 62.7  & 66.6  & 67.0    & 66.3 & 64.7  & 64.6    & 65.1  & 64.7  & 64.9    & 64.7  \\ \midrule
\multicolumn{14}{c}{\textbf{\textit{Translate-Test}}}                                                                                         \\ \midrule
\source                   & R   & 46.9 & 46.9 & 46.9 & 60.2 & 60.1 & 60.0 & 52.3 & 52.5 & 52.8 & 53.1 & 53.2 & 53.2     \\
\source                   & X   & 46.3 & 46.3 & 47.8 & 60.8 & 61.0 & 60.8 & 51.7 & 52.0 & 52.5 & 52.9 & 53.1 & 53.7    \\ \midrule
\multicolumn{14}{c}{\textbf{\textit{Roundtrip-Train-Test}}}                                                                                   \\ \midrule
\rndt                & R & 58.1 & 59.2 & 58.4 & 68.5 & 67.6 & 68.2 & 60.6 & 61.3 & 61.3 & 62.4 & 62.7 & 62.6    \\
\rndt                & X   & 58.9 & 59.3 & 59.3 & 69.7 & 69.7 & 69.3 & 60.7 & 60.4 & 60.6 & 63.1 & 63.1 & 63.1    \\
\rndtm              & R & 60.8 & 61.0 & 60.4 & 69.6 & 69.0 & 69.2 & 62.4 & 62.6 & 62.0 & 64.3 & 64.2 & 63.9    \\
\rndtm              & X   & 59.8 & 59.7 & 59.6 & 69.6 & 69.5 & 69.3 & 62.7 & 62.9 & 62.9 & 64.0 & 64.0 & 63.9     \\
\rndsens         & X   & 59.6 & 60.0 & 60.1 & 70.1 & 69.9 & 69.2 & 61.4 & 62.3 & 62.7 & 63.7 & 64.0 & 64.0     \\
\rndhens        & X   & 61.1 & 61.6 & 62.7 & 70.3 & 70.1 & 70.0 & 66.8 & 66.1 & 66.1 & 66.1 & 65.9 & 66.3    \\ \bottomrule
\end{tabular}
\caption{Results for translation-based XLT evaluated of AmNLI for languages supported by the translation model. Model selection is done on the best epoch based on source-language validation data (\src (I)), based on translated source-language validation data (\mtsrc (II)), and based on target-language validation data (\oracle (III)). We use XLM-R (X) and RoBERTa (R).}
\label{tab:supported_amnli}
\end{table*}

\begin{table*}[t!]
\scriptsize
\setlength{\tabcolsep}{2.2pt}
\centering
\begin{tabular}{@{}llcccccccccccccccccccccccc@{}}
\toprule
                    &         & \multicolumn{3}{c}{BZD} & \multicolumn{3}{c}{CNI} & \multicolumn{3}{c}{HCH} & \multicolumn{3}{c}{NAH} & \multicolumn{3}{c}{OTO} & \multicolumn{3}{c}{SHP} & \multicolumn{3}{c}{TAR} & \multicolumn{3}{c}{AVG} \\ \cmidrule{3-26} 
                    &         & I      & II     & III   & I      & II     & III   & I      & II     & III   & I      & II     & III   & I      & II     & III   & I      & II     & III   & I      & II     & III   & I      & II     & III   \\ \midrule
\multicolumn{26}{c}{\textit{Zero-Shot}}                                                                                                                                                                                                       \\ \midrule
\source                   & X   & 44.1   & 42.4   & 44.5  & 44.0   & 44.1   & 44.9  & 40.9   & 40.9   & 40.7  & 45.9   & 45.9   & 46.5  & 43.8   & 44.0   & 44.1  & 50.5   & 50.0   & 49.7  & 40.1   & 43.4   & 44.4  & 44.2   & 44.4   & 45.0  \\ \midrule
\multicolumn{26}{c}{\textbf{\textit{Translate-Train}}}                                                                                                                                                                                                 \\ \midrule
\jointh               & X   & 42.0   & 42.0   & 43.6  & 40.9   & 43.9   & 43.9  & 36.1   & 39.2   & 38.1  & 43.1   & 44.2   & 44.0  & 43.8   & 44.0   & 44.1  & 45.1   & 48.5   & 46.8  & 38.2   & 41.5   & 42.5  & 41.3   & 43.3   & 43.3  \\
\target                    & X   & 43.2   & 42.6   & 45.0  & 48.8   & 46.4   & 48.4  & 44.6   & 46.1   & 46.4  & 49.3   & 49.2   & 49.5  & 47.5   & 47.4   & 46.8  & 50.5   & 49.1   & 50.7  & 47.7   & 49.2   & 49.1  & 47.4   & 47.1   & 48.0  \\
\jointt                & X   & 44.9   & 44.4   & 45.7  & 47.6   & 47.5   & 48.8  & 44.8   & 45.0   & 45.7  & 48.4   & 48.4   & 48.6  & 47.8   & 47.8   & 48.0  & 51.0   & 48.0   & 51.0  & 47.7   & 48.5   & 49.0  & 47.5   & 47.1   & 48.1  \\
\seqt  & X   & 46.1   & 44.2   & 45.7  & 47.8   & 48.0   & 48.9  & 45.7   & 46.0   & 45.4  & 47.9   & 47.4   & 49.3  & 47.2   & 48.9   & 47.6  & 49.7   & 49.7   & 49.6  & 45.4   & 46.5   & 47.1  & 47.1   & 47.2   & 47.7  \\
\jointth             & X   & 44.5   & 44.4   & 44.9  & 46.8   & 47.0   & 47.6  & 44.7   & 44.8   & 45.6  & 49.2   & 50.1   & 48.9  & 47.3   & 48.1   & 47.4  & 48.1   & 47.8   & 49.0  & 49.4   & 49.1   & 49.6  & 47.2   & 47.3   & 47.6  \\
\tm                & X   & 45.9   & 44.9   & 46.2  & 46.1   & 46.1   & 45.6  & 45.0   & 44.8   & 45.1  & 49.6   & 49.1   & 48.6  & 46.3   & 46.9   & 45.5  & 48.5   & 48.9   & 48.8  & 47.2   & 46.6   & 49.5  & 46.9   & 46.8   & 47.0  \\
\jointtm               & X   & 45.5   & 45.5   & 46.1  & 45.5   & 46.6   & 46.7  & 44.4   & 44.9   & 44.6  & 48.1   & 47.7   & 48.7  & 46.9   & 46.9   & 46.1  & 49.5   & 49.2   & 50.2  & 45.6   & 45.8   & 46.4  & 46.5   & 46.7   & 47.0  \\
\seqtm& X   & 46.4   & 46.0   & 45.5  & 47.4   & 47.4   & 47.0  & 45.7   & 45.3   & 45.1  & 48.5   & 47.4   & 48.8  & 47.8   & 47.5   & 47.6  & 50.8   & 49.2   & 51.0  & 46.8   & 46.0   & 47.0  & 47.6   & 47.0   & 47.4  \\
\jointtmh            & X   & 45.2   & 45.3   & 46.9  & 45.8   & 46.5   & 46.5  & 45.2   & 45.1   & 45.0  & 48.2   & 48.6   & 50.1  & 47.0   & 47.4   & 47.1  & 50.0   & 50.1   & 50.7  & 46.3   & 46.2   & 47.7  & 46.8   & 47.0   & 47.7  \\ \midrule
\multicolumn{26}{c}{\textbf{\textit{Translate-Test}}}                                                                                                                                                                                                  \\ \midrule
\source                   & R & 35.8 & 36.0 & 35.6 & 32.9 & 33.7 & 33.1 & 36.5 & 36.1 & 36.9 & 39.5 & 40.1 & 39.6 & 38.4 & 38.2 & 37.3 & 38.5 & 38.8 & 39.2 & 33.8 & 34.4 & 33.5 & 36.5 & 36.8 & 36.5  \\
\source                   & X   & 35.3 & 35.1 & 36.1 & 35.8 & 36.4 & 36.4 & 37.0 & 37.1 & 36.3 & 38.8 & 39.2 & 38.7 & 39.4 & 39.4 & 38.0 & 41.4 & 40.9 & 40.8 & 33.8 & 34.3 & 33.9 & 37.4 & 37.5 & 37.2    \\ \midrule
\multicolumn{26}{c}{\textbf{\textit{Roundtrip-Train-Test}}}                                                                                                                                                                                            \\ \midrule
\rndt                & R & 36.4 & 36.7 & 36.8 & 36.5 & 36.9 & 36.7 & 37.3 & 36.5 & 37.2 & 39.8 & 39.8 & 39.5 & 41.5 & 40.6 & 40.6 & 42.7 & 42.1 & 41.5 & 34.5 & 34.7 & 34.0 & 38.4 & 38.2 & 38.0  \\
\rndt                & X   & 37.4 & 36.2 & 35.8 & 37.4 & 37.2 & 36.7 & 37.3 & 37.3 & 36.8 & 39.5 & 39.6 & 39.2 & 40.4 & 40.2 & 40.9 & 43.5 & 44.4 & 43.2 & 35.1 & 35.8 & 35.0 & 38.6 & 38.7 & 38.2   \\
\rndtm              & R & 37.1 & 37.5 & 37.1 & 38.9 & 39.3 & 37.9 & 38.4 & 37.9 & 38.6 & 39.4 & 39.4 & 40.2 & 40.9 & 40.8 & 41.8 & 41.9 & 41.7 & 43.3 & 35.3 & 34.7 & 34.5 & 38.8 & 38.7 & 39.0 \\
\rndtm              & X   & 37.1 & 36.8 & 37.2 & 39.0 & 38.8 & 37.8 & 39.6 & 39.4 & 39.5 & 41.1 & 40.4 & 41.0 & 39.3 & 39.8 & 39.4 & 43.2 & 42.8 & 42.9 & 34.8 & 34.8 & 34.9 & 39.1 & 39.0 & 38.9   \\
\rndsens         & X   & 37.0 & 37.0 & 36.8 & 38.7 & 39.0 & 38.2 & 39.2 & 38.6 & 39.4 & 41.3 & 40.2 & 40.3 & 39.4 & 38.6 & 41.2 & 43.5 & 42.5 & 43.2 & 34.8 & 34.5 & 35.1 & 39.1 & 38.6 & 39.2   \\
\rndhens        & X   & 41.1 & 40.7 & 41.4 & 39.1 & 38.9 & 39.2 & 39.9 & 40.7 & 40.5 & 43.7 & 43.3 & 44.9 & 40.2 & 40.9 & 42.2 & 46.6 & 47.2 & 46.7 & 37.4 & 38.3 & 37.6 & 41.1 & 41.4 & 41.8
\\ \bottomrule
\end{tabular}
\caption{Results for translation-based XLT evaluated of AmNLI for languages \textbf{not} supported by the translation model. Model selection is done on the best epoch based on source-language validation data (\src (I)), based on translated source-language validation data (\mtsrc (II)), and based on target-language validation data (\oracle (III)). We use XLM-R (X) and RoBERTa (R).}
\end{table*}

\begin{table*}[t!]
\scriptsize
\setlength{\tabcolsep}{2.2pt}
\centering
\begin{tabular}{@{}llcccccccccccccccccccccccc@{}}
\toprule
                    &         & \multicolumn{3}{c}{ACE} & \multicolumn{3}{c}{BAN} & \multicolumn{3}{c}{BJN} & \multicolumn{3}{c}{BUG} & \multicolumn{3}{c}{JAV} & \multicolumn{3}{c}{MIN} & \multicolumn{3}{c}{SUN} & \multicolumn{3}{c}{Avg} \\ \cmidrule{3-26} 
                    &         & I      & II     & III   & I      & II     & III   & I      & II     & III   & I      & II     & III   & I      & II     & III   & I      & II     & III   & I      & II     & III   & I      & II     & III   \\ \midrule
\multicolumn{26}{c}{\textit{Zero-Shot}}                                                                                                                                                                                                       \\ \midrule
\source                   & X   & 65.7   & 64.6   & 65.7  & 72.5   & 72.7   & 71.9  & 79.5   & 79.7   & 80.1  & 36.9   & 42.6   & 43.9  & 82.7   & 79.9   & 84.8  & 79.2   & 80.3   & 80.4  & 81.8   & 83.9   & 83.6  & 71.2   & 72.0   & 72.9  \\ \midrule
\multicolumn{26}{c}{\textbf{\textit{Translate-Train}}}                                                                                                                                                                                                \\ \midrule
\jointh               & X   & 67.0   & 68.0   & 68.8  & 72.0   & 72.5   & 73.0  & 80.4   & 80.4   & 80.6  & 39.6   & 44.1   & 43.5  & 80.7   & 83.7   & 86.0  & 77.2   & 79.1   & 78.9  & 81.3   & 80.8   & 81.0  & 71.2   & 72.7   & 73.1  \\
\target                    & X   & 74.1   & 74.4   & 75.3  & 73.2   & 75.5   & 74.0  & 83.4   & 82.7   & 82.1  & 62.2   & 64.6   & 64.7  & 86.1   & 86.0   & 88.9  & 82.2   & 83.2   & 83.1  & 83.6   & 83.4   & 84.3  & 77.8   & 78.6   & 78.9  \\
\jointt                & X   & 76.2   & 75.6   & 77.6  & 76.8   & 75.9   & 75.6  & 82.4   & 83.4   & 82.0  & 65.1   & 65.6   & 67.0  & 88.1   & 87.1   & 90.9  & 85.1   & 84.6   & 85.3  & 84.6   & 84.1   & 83.0  & 79.7   & 79.5   & 80.2  \\
\seqt  & X   & 74.6   & 75.0   & 75.6  & 76.3   & 77.0   & 77.0  & 81.9   & 82.1   & 82.6  & 64.7   & 62.9   & 65.4  & 86.9   & 87.2   & 89.7  & 83.5   & 84.3   & 82.9  & 84.1   & 83.9   & 83.6  & 78.9   & 78.9   & 79.5  \\
\jointth             & X   & 73.1   & 75.1   & 75.4  & 75.4   & 76.2   & 76.1  & 81.5   & 81.6   & 82.2  & 63.6   & 61.8   & 64.6  & 87.4   & 87.8   & 89.5  & 82.6   & 83.8   & 84.7  & 83.1   & 84.4   & 83.9  & 78.1   & 78.7   & 79.5  \\
\tm                & X   & 74.8   & 77.8   & 77.9  & 75.6   & 77.5   & 77.1  & 84.1   & 84.3   & 84.5  & 65.0   & 64.6   & 65.2  & 84.8   & 86.0   & 88.8  & 85.1   & 84.0   & 84.3  & 83.6   & 84.4   & 84.6  & 79.0   & 79.8   & 80.3  \\
\jointtm               & X   & 77.7   & 77.8   & 76.4  & 77.4   & 77.3   & 78.5  & 86.1   & 84.8   & 86.0  & 65.1   & 66.2   & 66.8  & 86.5   & 84.4   & 88.3  & 86.5   & 85.8   & 86.2  & 86.5   & 86.4   & 86.5  & 80.8   & 80.4   & 81.2  \\
\seqtm& X   & 75.3   & 76.7   & 75.4  & 77.1   & 78.5   & 78.0  & 84.2   & 83.6   & 85.0  & 64.0   & 66.8   & 67.7  & 84.0   & 83.2   & 88.2  & 84.3   & 84.4   & 84.4  & 85.6   & 85.7   & 83.5  & 79.2   & 79.8   & 80.3  \\
\jointtmh            & X   & 76.8   & 78.2   & 77.8  & 76.5   & 78.0   & 77.1  & 84.0   & 84.1   & 84.9  & 65.6   & 66.9   & 66.0  & 81.8   & 84.9   & 88.5  & 83.5   & 83.9   & 85.1  & 85.5   & 85.1   & 85.4  & 79.1   & 80.2   & 80.7  \\ \midrule
\multicolumn{26}{c}{\textbf{\textit{Translate-Test}}}                                                                                                                                                                                                  \\ \midrule
\source                   & R &  77.3 & 75.5 & 77.5 & 74.1 & 75.5 & 75.8 & 82.2 & 79.6 & 82.0 & 69.5 & 71.8 & 72.3 & 85.8 & 84.3 & 85.5 & 81.9 & 82.0 & 82.9 & 84.8 & 84.3 & 84.8 & 79.4 & 79.0 & 80.1   \\
\source                   & X   & 78.8 & 77.9 & 78.5 & 77.2 & 77.4 & 78.8 & 83.6 & 83.3 & 82.3 & 71.7 & 70.1 & 74.5 & 85.5 & 86.1 & 85.8 & 83.4 & 83.6 & 84.6 & 86.1 & 86.3 & 85.8 & 80.9 & 80.7 & 81.5    \\ \midrule
\multicolumn{26}{c}{\textbf{\textit{Roundtrip-Train-Test}}}                                                                                                                                                                                            \\ \midrule
\rndt                & R & 79.5 & 79.3 & 79.1 & 76.1 & 77.9 & 77.8 & 82.8 & 82.4 & 82.2 & 74.5 & 74.3 & 73.7 & 85.7 & 83.7 & 85.0 & 85.6 & 84.3 & 84.3 & 84.6 & 84.7 & 85.3 & 81.2 & 81.0 & 81.0 \\
\rndt                & X   & 78.3 & 79.4 & 79.7 & 78.8 & 78.1 & 77.1 & 83.9 & 83.8 & 84.1 & 73.1 & 74.1 & 75.3 & 86.5 & 86.8 & 86.4 & 84.9 & 85.5 & 85.9 & 85.6 & 84.9 & 85.7 & 81.6 & 81.8 & 82.0    \\
\rndtm              & R & 78.6 & 77.8 & 78.2 & 77.8 & 79.3 & 80.1 & 83.6 & 83.6 & 83.4 & 73.8 & 73.4 & 74.6 & 85.8 & 85.3 & 86.0 & 83.9 & 84.2 & 84.2 & 83.7 & 83.6 & 83.5 & 81.0 & 81.0 & 81.4  \\
\rndtm              & X   & 78.8 & 78.6 & 79.8 & 79.6 & 78.0 & 80.3 & 85.2 & 84.8 & 85.0 & 74.5 & 75.1 & 75.6 & 86.9 & 87.1 & 86.6 & 84.7 & 84.3 & 84.8 & 85.0 & 85.1 & 84.8 & 82.1 & 81.9 & 82.4 \\
\rndsens         & X   & 79.8 & 79.1 & 80.2 & 80.2 & 80.0 & 80.5 & 86.5 & 86.5 & 86.3 & 74.8 & 75.8 & 75.7 & 87.5 & 87.3 & 86.6 & 85.3 & 85.8 & 85.3 & 85.8 & 85.7 & 84.2 & 82.8 & 82.9 & 82.7   \\
\rndhens        & X   & 83.2 & 83.5 & 83.2 & 82.2 & 81.6 & 82.4 & 86.0 & 85.7 & 85.1 & 75.2 & 75.5 & 74.6 & 88.0 & 88.0 & 87.0 & 86.5 & 86.1 & 86.7 & 86.5 & 86.9 & 86.0 & 83.9 & 83.9 & 83.6
\\ \bottomrule
\end{tabular}
\caption{Results for translation-based XLT evaluated of NusaX for languages supported by the translation model. Model selection is done on the best epoch based on source-language validation data (\src (I)), based on translated source-language validation data (\mtsrc (II)), and based on target-language validation data (\oracle (III)). We use XLM-R (X) and RoBERTa (R).}
\end{table*}

\begin{table*}[t!]
\small
\setlength{\tabcolsep}{7.8pt}
\centering
\begin{tabular}{@{}llcccccccccccc@{}}
\toprule
                    &         & \multicolumn{3}{c}{BBC} & \multicolumn{3}{c}{MAD} & \multicolumn{3}{c}{NIJ} & \multicolumn{3}{c}{Avg} \\ \cmidrule{3-14}
                    &         & I      & II     & III   & I      & II     & III   & I      & II     & III   & I      & II     & III   \\ \midrule
\multicolumn{14}{c}{\textit{Zero-Shot}}                                                                                               \\ \midrule
\source                   & X   & 41.4   & 45.5   & 45.9  & 65.5   & 64.8   & 67.4  & 66.6   & 65.7   & 67.2  & 57.8   & 58.7   & 60.2  \\ \midrule
\multicolumn{14}{c}{\textbf{\textit{Translate-Train}}}                                                                                         \\ \midrule
\jointh               & X   & 42.7   & 46.5   & 45.3  & 65.1   & 62.8   & 68.5  & 62.7   & 62.1   & 65.6  & 56.8   & 57.1   & 59.8  \\
\target                    & X   & 60.6   & 60.9   & 62.2  & 70.1   & 69.6   & 73.1  & 66.1   & 67.4   & 69.9  & 65.6   & 65.9   & 68.4  \\
\jointt                & X   & 61.2   & 62.2   & 64.0  & 72.4   & 72.4   & 71.9  & 69.0   & 69.7   & 68.6  & 67.5   & 68.1   & 68.2  \\
\seqt  & X   & 63.8   & 62.7   & 66.1  & 70.7   & 69.5   & 70.7  & 69.6   & 69.2   & 70.1  & 68.0   & 67.1   & 68.9  \\
\jointth             & X   & 62.2   & 63.2   & 61.7  & 71.7   & 72.1   & 72.4  & 71.5   & 68.9   & 71.6  & 68.5   & 68.1   & 68.6  \\
\tm                & X   & 65.8   & 67.8   & 66.8  & 76.8   & 75.6   & 78.9  & 74.8   & 74.5   & 76.2  & 72.5   & 72.6   & 74.0  \\
\jointtm               & X   & 66.3   & 67.9   & 65.2  & 78.2   & 76.6   & 77.8  & 77.5   & 75.7   & 77.8  & 74.0   & 73.4   & 73.6  \\
\seqtm& X   & 68.0   & 68.3   & 65.6  & 77.8   & 77.9   & 78.0  & 76.1   & 77.2   & 78.4  & 74.0   & 74.5   & 74.0  \\ 
\jointtmh            & X   & 65.1   & 66.9   & 64.0  & 76.7   & 75.5   & 77.0  & 75.2   & 75.3   & 76.8  & 72.3   & 72.6   & 72.6  \\ \midrule
\multicolumn{14}{c}{\textbf{\textit{Translate-Test}}}                                                                                          \\ \midrule
\source                   & R & 42.6 & 47.8 & 49.2 & 56.4 & 56.4 & 58.8 & 64.3 & 63.7 & 65.8 & 54.4 & 56.0 & 57.9     \\
\source                   & X   & 40.4 & 38.5 & 55.1 & 60.9 & 59.8 & 63.6 & 63.4 & 62.1 & 65.8 & 54.9 & 53.5 & 61.5     \\ \midrule
\multicolumn{14}{c}{\textbf{\textit{Roundtrip-Train-Test}}}                                                                                    \\ \midrule
\rndt                & R & 49.6 & 45.5 & 50.2 & 55.1 & 56.1 & 58.1 & 60.9 & 62.0 & 63.1 & 55.2 & 54.5 & 57.1     \\
\rndt                & X   & 44.0 & 46.6 & 54.6 & 62.5 & 62.2 & 64.3 & 64.6 & 65.2 & 64.3 & 57.0 & 58.0 & 61.1    \\
\rndtm              & R & 51.5 & 50.5 & 52.5 & 61.7 & 60.8 & 61.6 & 68.3 & 66.4 & 68.3 & 60.5 & 59.2 & 60.8     \\
\rndtm              & X   & 47.2 & 54.2 & 55.3 & 62.7 & 64.6 & 66.7 & 67.3 & 68.5 & 67.0 & 59.1 & 62.4 & 63.0     \\
\rndsens         & X   & 49.4 & 54.1 & 56.4 & 65.7 & 68.0 & 68.5 & 68.3 & 69.5 & 69.6 & 61.1 & 63.9 & 64.8       \\
\rndhens        & X   & 51.9 & 56.9 & 58.1 & 69.8 & 68.8 & 70.9 & 73.2 & 72.5 & 72.7 & 65.0 & 66.1 & 67.2    \\ \bottomrule
\end{tabular}
\caption{Results for translation-based XLT evaluated of NusaX for languages \textbf{not} supported by the translation model. Model selection is done on the best epoch based on source-language validation data (\src (I)), based on translated source-language validation data (\mtsrc (II)), and based on target-language validation data (\oracle (III)). We use XLM-R (X) and RoBERTa (R).}
\end{table*}

\begin{table*}[t!]
\scriptsize
\setlength{\tabcolsep}{3.4pt}
\centering
\begin{tabular}{@{}llccccccccccccccccccc@{}}
\toprule
                    &   & BAM  & EWE  & FON  & HAU  & IBO  & KIN  & LUG  & LUO  & MOS  & NYA  & SNA  & SWA  & TSN  & TWI  & WOL  & XHO  & YOR  & ZUL  & Avg  \\ \midrule
\multicolumn{21}{c}{\textit{Zero-Shot}}                                                                                                                      \\ \midrule
\source                   & X & 36.9 & 67.9 & 46.8 & 73.4 & 48.0 & 42.0 & 58.6 & 37.7 & 47.6 & 47.4 & 35.8 & 85.5 & 48.1 & 43.3 & 48.2 & 22.8 & 31.1 & 41.1 & 47.9 \\ \midrule
\multicolumn{21}{c}{\textbf{\textit{Translate-Train}}}                                                                                                                \\ \midrule
\jointh               & X & 35.8 & 70.4 & 50.5 & 72.5 & 54.6 & 43.3 & 63.9 & 40.9 & 50.6 & 53.3 & 55.4 & 81.9 & 52.5 & 42.7 & 52.3 & 56.9 & 33.6 & 59.1 & 53.9 \\
\target                    & X & 51.3 & 72.6 & 64.5 & 71.4 & 65.8 & 53.5 & 69.7 & 47.7 & 53.9 & 63.9 & 65.6 & 76.3 & 68.0 & 60.3 & 58.8 & 68.7 & 36.4 & 69.0 & 62.1 \\
\jointt                & X & 48.9 & 75.4 & 65.9 & 72.3 & 68.1 & 54.7 & 74.3 & 50.1 & 57.0 & 68.0 & 69.9 & 77.4 & 68.7 & 61.5 & 61.7 & 70.0 & 38.0 & 71.4 & 64.1 \\
\seqt  & X & 51.0 & 71.4 & 65.7 & 72.1 & 68.3 & 54.2 & 73.2 & 48.8 & 56.1 & 65.0 & 67.4 & 76.2 & 69.7 & 60.1 & 56.9 & 69.1 & 38.1 & 70.9 & 63.0 \\
\jointth             & X & 47.9 & 71.8 & 67.2 & 71.5 & 70.2 & 54.4 & 73.3 & 48.6 & 54.5 & 66.6 & 68.1 & 76.1 & 68.0 & 61.0 & 58.0 & 68.5 & 38.0 & 68.7 & 62.9 \\
\tm                & X & 43.8 & 65.1 & 60.7 & 69.2 & 63.7 & 51.1 & 66.2 & 47.1 & 45.2 & 57.0 & 62.1 & 75.2 & 58.2 & 58.9 & 48.4 & 58.1 & 36.2 & 58.6 & 56.9 \\
\jointtm               & X & 44.1 & 65.5 & 58.1 & 70.1 & 61.8 & 53.2 & 66.7 & 45.6 & 46.7 & 56.6 & 60.7 & 76.1 & 62.4 & 59.7 & 46.8 & 61.2 & 35.9 & 62.8 & 57.4 \\
\seqtm& X & 48.3 & 68.0 & 63.7 & 69.8 & 64.8 & 54.0 & 67.0 & 48.4 & 50.1 & 58.6 & 61.2 & 75.9 & 61.2 & 60.1 & 52.5 & 62.0 & 38.7 & 62.5 & 59.3 \\
\jointtmh            & X & 45.7 & 65.4 & 64.3 & 69.0 & 64.9 & 52.7 & 65.2 & 46.5 & 49.2 & 56.9 & 60.7 & 75.3 & 57.9 & 58.6 & 53.9 & 60.4 & 36.7 & 61.7 & 58.0 \\ \midrule
\multicolumn{21}{c}{\textbf{\textit{Translate-Test}}}                                                                                                                 \\ \midrule
\source                   & R & 39.9 & 61.3 & 56.4 & 58.0 & 55.8 & 51.6 & 68.1 & 45.5 & 39.6 & 63.7 & 58.5 & 62.0 & 60.1 & 56.8 & 49.7 & 58.0 & 43.9 & 57.0 & 54.8 \\
\source                   & X & 39.4 & 61.5 & 56.3 & 57.8 & 54.9 & 50.5 & 67.9 & 43.2 & 39.1 & 63.1 & 58.0 & 61.6 & 57.9 & 55.2 & 49.9 & 57.6 & 43.4 & 56.7 & 54.1 \\ \midrule
\multicolumn{21}{c}{\textbf{\textit{Roundtrip-Train-Test}}}                                                                                                           \\ \midrule
\rndt                & R & 39.7 & 61.2 & 57.2 & 58.3 & 60.6 & 49.9 & 65.6 & 44.0 & 37.6 & 63.8 & 57.8 & 62.2 & 59.8 & 57.2 & 50.9 & 55.9 & 45.2 & 57.4 & 54.7 \\
\rndt                & X & 39.0 & 60.0 & 57.2 & 57.8 & 58.2 & 50.6 & 65.1 & 42.6 & 36.4 & 62.5 & 57.0 & 61.8 & 57.6 & 55.2 & 50.1 & 55.0 & 44.3 & 56.5 & 53.7 \\
\rndtm              & R & 40.0 & 57.9 & 55.0 & 58.3 & 59.8 & 49.6 & 63.7 & 43.0 & 35.5 & 62.3 & 55.3 & 62.7 & 59.5 & 55.6 & 50.1 & 54.5 & 43.7 & 55.4 & 53.4 \\
\rndtm              & X & 39.1 & 59.0 & 55.8 & 58.4 & 59.6 & 49.3 & 65.1 & 41.1 & 36.9 & 61.1 & 55.3 & 62.4 & 58.2 & 56.0 & 50.2 & 53.5 & 43.0 & 55.7 & 53.3 \\ 
\rndsens            & X & 39.9 & 59.2 & 56.2 & 58.0 & 60.3 & 50.2 & 64.8 & 41.5 & 38.1 & 61.7 & 56.0 & 62.4 & 57.0 & 56.5 & 50.9 & 54.1 & 44.2 & 55.8 & 53.7 \\
\rndhens            & X  & 33.8 & 50.6 & 46.7 & 47.6 & 50.4 & 42.1 & 53.7 & 34.7 & 34.7 & 53.1 & 50.2 & 54.1 & 48.3 & 47.0 & 44.3 & 47.9 & 38.6 & 46.6 & 45.8 \\
\bottomrule
\end{tabular}
\caption{Results for translation-based XLT evaluated of Masakha for languages supported by the translation model. Model selection is done on the best epoch based on source-language validation data (\src). We use XLM-R (X) and RoBERTa (R).}
\end{table*}

\begin{table*}[t!]
\small
\setlength{\tabcolsep}{3.4pt}
\scriptsize
\begin{tabular}{@{}llccccccccccccccccccc@{}}
\toprule
                    &   & BAM  & EWE  & FON  & HAU  & IBO  & KIN  & LUG  & LUO  & MOS  & NYA  & SNA  & SWA  & TSN  & TWI  & WOL  & XHO  & YOR  & ZUL  & Avg  \\ \midrule
\multicolumn{21}{c}{\textit{Zero-Shot}}                                                                                                                      \\ \midrule
\source                   & X & 38.9 & 69.1 & 49.4 & 73.2 & 50.6 & 43.3 & 62.4 & 38.4 & 49.8 & 49.0 & 35.7 & 85.3 & 49.6 & 45.2 & 50.9 & 22.6 & 32.4 & 41.3 & 49.3 \\ \midrule
\multicolumn{21}{c}{\textbf{\textit{Translate-Train}}}                                                                                                                \\ \midrule
\jointh               & X & 38.7 & 72.4 & 54.4 & 72.6 & 58.5 & 46.0 & 65.5 & 40.6 & 51.9 & 54.4 & 54.6 & 82.0 & 52.9 & 47.7 & 51.6 & 57.3 & 33.7 & 59.4 & 55.2 \\
\target                    & X & 50.0 & 74.4 & 65.0 & 71.2 & 65.7 & 53.8 & 73.0 & 48.4 & 55.0 & 64.6 & 66.3 & 76.4 & 68.6 & 58.7 & 58.8 & 68.2 & 37.1 & 70.2 & 62.5 \\
\jointt                & X & 50.6 & 75.5 & 66.0 & 72.2 & 68.6 & 55.3 & 75.0 & 50.0 & 55.6 & 67.5 & 69.2 & 77.7 & 69.5 & 61.8 & 60.6 & 69.3 & 38.6 & 72.1 & 64.2 \\
\seqt  & X & 50.8 & 73.6 & 66.0 & 72.0 & 68.5 & 56.0 & 74.9 & 50.0 & 56.1 & 67.0 & 70.0 & 76.9 & 69.9 & 61.8 & 61.0 & 69.5 & 38.4 & 71.3 & 64.1 \\
\jointth             & X & 50.9 & 72.0 & 67.5 & 71.6 & 69.7 & 54.1 & 73.9 & 49.1 & 54.1 & 67.5 & 69.3 & 76.8 & 68.7 & 61.5 & 57.9 & 68.8 & 39.1 & 70.2 & 63.5 \\
\tm                & X & 46.5 & 64.4 & 59.0 & 69.4 & 64.1 & 51.4 & 65.4 & 45.3 & 46.7 & 56.9 & 60.3 & 74.6 & 59.6 & 58.3 & 52.4 & 59.6 & 36.6 & 60.0 & 57.2 \\
\jointtm               & X & 44.7 & 65.5 & 59.4 & 68.8 & 66.2 & 51.7 & 63.3 & 46.8 & 47.4 & 56.7 & 60.3 & 74.8 & 59.9 & 57.4 & 53.5 & 59.8 & 36.0 & 61.8 & 57.4 \\
\seqtm& X & 45.1 & 64.6 & 62.9 & 69.6 & 65.1 & 51.6 & 67.3 & 46.1 & 46.4 & 55.6 & 60.4 & 73.8 & 59.1 & 60.1 & 51.1 & 61.3 & 34.9 & 61.1 & 57.6 \\
\jointtmh            & X & 45.0 & 64.0 & 59.5 & 68.4 & 65.5 & 51.0 & 62.3 & 47.1 & 48.0 & 56.5 & 60.3 & 74.6 & 59.1 & 57.9 & 51.7 & 59.3 & 33.6 & 61.0 & 56.9 \\ 
\midrule
\multicolumn{21}{c}{\textbf{\textit{Translate-Test}}}                                                                                                                 \\ \midrule
\source                   & R & 39.9 & 61.4 & 56.3 & 58.0 & 55.9 & 51.4 & 67.8 & 44.8 & 39.7 & 63.7 & 58.5 & 62.0 & 60.0 & 56.3 & 49.8 & 57.6 & 44.1 & 57.4 & 54.7 \\
\source                   & X & 39.2 & 61.2 & 55.6 & 57.8 & 54.8 & 50.6 & 67.7 & 43.0 & 39.1 & 63.3 & 57.8 & 61.7 & 57.8 & 54.5 & 49.7 & 57.5 & 43.0 & 56.6 & 53.9 \\ \midrule
\multicolumn{21}{c}{\textbf{\textit{Roundtrip-Train-Test}}}                                                                                                           \\ \midrule
\rndt                & R & 39.7 & 61.1 & 56.9 & 58.4 & 61.2 & 50.2 & 65.9 & 44.5 & 37.7 & 63.5 & 57.8 & 62.4 & 59.8 & 56.9 & 51.4 & 55.9 & 45.5 & 57.0 & 54.8 \\
\rndt                & X & 39.7 & 59.6 & 56.5 & 58.0 & 58.6 & 50.7 & 66.0 & 42.5 & 36.4 & 62.7 & 57.1 & 61.9 & 57.0 & 54.1 & 50.7 & 55.1 & 44.3 & 56.0 & 53.7 \\
\rndtm              & R & 39.1 & 58.4 & 56.6 & 58.1 & 60.2 & 49.1 & 62.3 & 42.0 & 35.1 & 62.3 & 56.2 & 62.2 & 59.8 & 57.2 & 50.3 & 54.5 & 43.2 & 55.7 & 53.5 \\
\rndtm              & X & 40.4 & 57.2 & 55.4 & 58.1 & 61.2 & 49.0 & 62.5 & 42.0 & 35.9 & 61.5 & 54.7 & 62.1 & 57.8 & 55.7 & 49.7 & 50.9 & 42.6 & 54.2 & 52.8 \\ 
\rndsens            & X & 40.2 & 58.8 & 55.9 & 58.3 & 60.7 & 50.0 & 64.7 & 40.8 & 36.8 & 61.8 & 56.0 & 62.7 & 57.1 & 56.2 & 51.2 & 53.7 & 43.7 & 55.8 & 53.6 \\
\rndhens            & X  & 33.9 & 50.6 & 47.1 & 47.9 & 50.3 & 41.9 & 53.3 & 34.2 & 34.5 & 53.3 & 49.8 & 54.7 & 48.3 & 47.4 & 44.5 & 47.8 & 37.8 & 46.9 & 45.8 \\
\bottomrule
\end{tabular}
\caption{Results for translation-based XLT evaluated of Masakha for languages supported by the translation model. Model selection is done on the best epoch based on translated source-language validation data (\mtsrc). We use XLM-R (X) and RoBERTa (R).}
\end{table*}

\begin{table*}[t!]
\scriptsize
\setlength{\tabcolsep}{3.4pt}
\centering
\begin{tabular}{@{}llccccccccccccccccccc@{}}
\toprule
                    &   & BAM  & EWE  & FON  & HAU  & IBO  & KIN  & LUG  & LUO  & MOS  & NYA  & SNA  & SWA  & TSN  & TWI  & WOL  & XHO  & YOR  & ZUL  & Avg  \\ \midrule
\multicolumn{21}{c}{\textit{Zero-Shot}}                                                                                                                      \\ \midrule
\source                   & X & 39.6 & 70.8 & 50.9 & 73.4 & 52.9 & 43.5 & 64.7 & 39.3 & 49.6 & 51.6 & 40.6 & 85.5 & 52.7 & 46.0 & 51.6 & 22.2 & 34.4 & 41.9 & 50.6 \\ \midrule
\multicolumn{21}{c}{\textbf{\textit{Translate-Train}}}                                                                                                                \\ \midrule
\jointh               & X & 40.3 & 72.3 & 56.2 & 72.9 & 60.9 & 46.4 & 66.0 & 39.9 & 53.9 & 54.1 & 55.9 & 84.0 & 53.4 & 49.5 & 53.8 & 57.2 & 35.2 & 60.9 & 56.3 \\
\target                   & X & 52.0 & 75.5 & 64.7 & 71.3 & 66.8 & 54.5 & 75.0 & 49.5 & 59.3 & 64.6 & 67.9 & 76.6 & 67.8 & 61.8 & 59.5 & 67.9 & 37.7 & 69.8 & 63.4 \\
\jointt                & X & 54.6 & 77.1 & 67.1 & 72.6 & 69.9 & 56.8 & 76.5 & 50.9 & 58.5 & 68.3 & 70.2 & 79.2 & 69.8 & 62.4 & 61.8 & 70.1 & 40.2 & 72.9 & 65.5 \\
\seqt   & X & 52.0 & 75.5 & 66.8 & 72.8 & 69.5 & 56.8 & 76.5 & 49.3 & 59.1 & 68.0 & 70.1 & 77.9 & 69.8 & 61.3 & 61.6 & 69.9 & 39.7 & 71.7 & 64.9 \\
\jointth             & X & 52.9 & 74.4 & 68.1 & 73.0 & 70.2 & 55.0 & 74.7 & 49.1 & 57.8 & 69.1 & 70.0 & 77.8 & 68.5 & 61.9 & 60.2 & 69.5 & 40.1 & 69.6 & 64.5 \\
\tm                & X & 49.1 & 71.7 & 63.4 & 71.3 & 66.2 & 54.9 & 66.2 & 47.6 & 49.3 & 58.4 & 63.0 & 76.9 & 62.9 & 57.7 & 54.9 & 63.4 & 37.8 & 63.7 & 59.9 \\
\jointtm               & X & 49.0 & 70.0 & 61.2 & 71.0 & 67.3 & 53.5 & 69.4 & 47.1 & 50.4 & 59.1 & 62.7 & 76.8 & 62.3 & 58.3 & 55.9 & 62.6 & 39.1 & 64.7 & 60.0 \\
\seqtm& X & 48.7 & 70.3 & 64.7 & 70.8 & 66.6 & 55.1 & 69.6 & 49.4 & 50.4 & 59.9 & 62.2 & 76.0 & 62.1 & 59.2 & 52.6 & 63.0 & 38.2 & 63.7 & 60.1 \\
\jointtmh            & X & 48.6 & 68.3 & 64.5 & 70.9 & 68.2 & 53.9 & 68.9 & 45.8 & 49.3 & 59.3 & 62.9 & 76.6 & 63.3 & 61.8 & 55.5 & 63.0 & 39.2 & 63.1 & 60.2 \\ \midrule
\multicolumn{21}{c}{\textbf{\textit{Translate-Test}}}                                                                                                                 \\ \midrule
\source                   & R & 39.7 & 61.4 & 56.6 & 58.0 & 56.5 & 51.6 & 67.9 & 45.0 & 39.7 & 63.6 & 58.4 & 61.8 & 59.6 & 56.9 & 49.8 & 57.8 & 44.3 & 57.0 & 54.7
 \\
\source                   & X & 39.8 & 61.1 & 56.1 & 57.8 & 55.1 & 50.7 & 67.9 & 42.5 & 38.9 & 63.3 & 57.7 & 61.6 & 57.8 & 54.9 & 49.6 & 57.3 & 43.0 & 56.7 & 54.0
 \\ \midrule
\multicolumn{21}{c}{\textbf{\textit{Roundtrip-Train-Test}}}                                                                                                           \\ \midrule
\rndt                & R & 40.6 & 60.3 & 56.5 & 58.3 & 61.1 & 51.1 & 66.9 & 43.4 & 38.0 & 63.6 & 57.8 & 62.6 & 60.3 & 57.0 & 51.7 & 55.7 & 45.4 & 56.0 & 54.8 \\
\rndt                & X & 40.4 & 60.4 & 57.3 & 58.2 & 58.3 & 50.6 & 66.9 & 41.8 & 37.1 & 63.0 & 56.9 & 62.4 & 55.7 & 56.8 & 51.0 & 55.2 & 44.6 & 56.9 & 54.1 \\
\rndtm              & R & 40.9 & 59.1 & 55.9 & 57.9 & 61.6 & 50.1 & 65.1 & 42.7 & 36.5 & 62.4 & 55.5 & 63.7 & 59.6 & 57.2 & 50.7 & 53.7 & 44.1 & 55.7 & 54.0 \\
\rndtm              & X & 40.7 & 59.3 & 55.4 & 58.1 & 61.5 & 49.4 & 63.9 & 40.2 & 36.6 & 61.2 & 55.0 & 63.1 & 58.2 & 56.3 & 49.9 & 50.0 & 44.0 & 54.9 & 53.2 \\ 
\rndsens            & X & 40.5 & 59.8 & 55.9 & 58.2 & 61.0 & 50.6 & 65.5 & 41.7 & 38.1 & 62.0 & 56.1 & 63.3 & 57.7 & 57.3 & 51.0 & 51.7 & 44.7 & 55.2 & 53.9 \\
\rndhens            & X  & 35.4 & 52.4 & 48.3 & 48.4 & 51.8 & 43.3 & 57.1 & 35.4 & 35.5 & 53.4 & 50.6 & 55.3 & 49.3 & 49.1 & 45.4 & 48.1 & 40.2 & 49.4 & 47.1 \\
\bottomrule
\end{tabular}
\caption{Results for translation-based XLT evaluated of Masakha for languages supported by the translation model. Model selection is done on the best epoch based on target-language validation data (\oracle). We use XLM-R (X) and RoBERTa (R).}
\end{table*}

\begin{table*}[t!]
\small
\setlength{\tabcolsep}{12.5pt}
\centering
\begin{tabular}{@{}llccccccccc@{}}
\toprule
                    &   & \multicolumn{3}{c}{BBJ} & \multicolumn{3}{c}{PCM} & \multicolumn{3}{c}{Avg} \\ \midrule
                    &   & I      & II     & III   & I      & II     & III   & I      & II     & III   \\ \midrule
\multicolumn{11}{c}{\textit{Zero-Shot}}                                                               \\ \midrule
\source                   & X & 41.9   & 42.0   & 45.4  & 78.5   & 78.3   & 78.2  & 60.2   & 60.1   & 61.8  \\ \midrule
\multicolumn{11}{c}{\textbf{\textit{Translate-Train}}}                                                         \\ \midrule
\jointh                & X & 45.8   & 45.7   & 44.6  & 77.2   & 77.3   & 76.5  & 61.5   & 61.5   & 60.6  \\
\target                    & X & 43.2   & 41.8   & 44.1  & 75.0   & 75.7   & 75.9  & 59.1   & 58.7   & 60.0  \\
\jointt                & X & 46.3   & 46.5   & 48.7  & 77.3   & 77.2   & 77.6  & 61.8   & 61.8   & 63.2  \\
\seqt  & X & 46.0   & 46.7   & 47.5  & 76.3   & 76.4   & 77.1  & 61.2   & 61.5   & 62.3  \\
\jointth              & X & 42.0   & 44.8   & 46.3  & 77.0   & 77.0   & 77.4  & 59.5   & 60.9   & 61.9  \\
\tm                & X & 48.4   & 48.1   & 49.9  & 72.6   & 73.7   & 73.0  & 60.5   & 60.9   & 61.4  \\
\jointtm               & X & 47.9   & 47.0   & 51.0  & 74.0   & 72.9   & 75.8  & 61.0   & 60.0   & 63.4  \\
\seqtm& X & 50.0   & 46.7   & 51.0  & 73.9   & 72.7   & 73.5  & 61.9   & 59.7   & 62.2  \\ 
\jointtmh            & X & 48.4   & 47.3   & 49.8  & 73.4   & 72.8   & 75.1  & 60.9   & 60.1   & 62.4  \\ \midrule
\multicolumn{11}{c}{\textbf{\textit{Translate-Test}}}                                                          \\ \midrule
\source                   & R & 31.8 & 31.7 & 32.4 & 64.4 & 64.3 & 64.4 & 48.1 & 48.0 & 48.4     \\
\source                   & X & 30.5 & 30.3 & 32.1 & 62.6 & 62.4 & 62.5 & 46.6 & 46.4 & 47.3
     \\ \midrule
\multicolumn{11}{c}{\textbf{\textit{Roundtrip-Train-Test}}}                                                    \\ \midrule
\rndt                & R & 30.4 & 29.7 & 31.7 & 61.9 & 62.1 & 62.9 & 46.2 & 45.9 & 47.3     \\
\rndt                & X & 30.6 & 30.6 & 32.3 & 60.0 & 59.5 & 60.3 & 45.3 & 45.1 & 46.3     \\
\rndtm              & R & 30.8 & 30.8 & 34.1 & 59.3 & 58.7 & 59.4 & 45.0 & 44.7 & 46.8     \\
\rndtm              & X & 30.4 & 31.0 & 32.4 & 57.6 & 56.4 & 58.2 & 44.0 & 43.7 & 45.3   \\ 
\rndsens            & X & 34.5 & 35.4 & 35.4 & 57.9 & 57.3 & 58.9 & 46.2 & 46.3 & 47.2 \\
\rndhens            & X  & 35.4 & 35.1 & 37.1 & 50.2 & 49.3 & 50.3 & 42.8 & 42.2 & 43.7
 \\
\bottomrule
\end{tabular}
\caption{Results for translation-based XLT evaluated of Masakha for languages \textbf{not} supported by the translation model. Model selection is done on the best epoch based on source-language validation data (\src (I)), based on translated source-language validation data (\mtsrc (II)), and based on target-language validation data (\oracle (III)). We use XLM-R (X) and RoBERTa (R).}
\end{table*}

\end{document}